\documentclass{article}

\usepackage{microtype}
\usepackage{graphicx}
\usepackage{booktabs} 

\usepackage{subcaption}

\usepackage{hyperref}
\usepackage{multirow}

\usepackage[accepted]{icml2023}

\usepackage{amsmath}
\usepackage{amssymb}
\usepackage{mathtools}
\usepackage{amsthm}
\usepackage{bm}

\usepackage{ulem}
\usepackage{cancel}
\usepackage[capitalize,noabbrev]{cleveref}

\theoremstyle{plain}
\newtheorem{theorem}{Theorem}[section]
\newtheorem{proposition}[theorem]{Proposition}
\newtheorem{lemma}[theorem]{Lemma}

\theoremstyle{definition}

\theoremstyle{remark}

\newcommand{\se}[1]{\textcolor{magenta}{}}

\newcommand{\nm}[1]{{\textcolor{black}{#1}}}

\newcommand{\ks}[1]{{\textcolor{black}{#1}}}

\newcommand{\md}[1]{{\textcolor{black}{#1}}}

\newcommand{\camred}[1]{{\textcolor{black}{#1}}}

\usepackage[textsize=tiny]{todonotes}

\newcommand{\sT}{\mathsf{T}}

\icmltitlerunning{\hfill GibbsDDRM: A Partially Collapsed Gibbs Sampler for Solving Blind Inverse Problems with Denoising Diffusion Restoration \hfill }
\begin{document}

\twocolumn[
\icmltitle{GibbsDDRM: A Partially Collapsed Gibbs Sampler for \newline Solving Blind Inverse Problems with Denoising Diffusion Restoration}



\icmlsetsymbol{equal}{*}

\begin{icmlauthorlist}
\icmlauthor{Naoki Murata}{sonyai}
\icmlauthor{Koichi Saito}{sonyai}
\icmlauthor{Chieh-Hsin Lai}{sonyai}
\icmlauthor{Yuhta Takida}{sonyai}
\icmlauthor{Toshimitsu Uesaka}{sonyai}
\icmlauthor{Yuki Mitsufuji}{sonyai,sony}
\icmlauthor{Stefano Ermon}{stu}
\end{icmlauthorlist}

\icmlaffiliation{sonyai}{Sony AI, Tokyo, Japan}
\icmlaffiliation{sony}{Sony Group Corporation, Tokyo, Japan}
\icmlaffiliation{stu}{Department of Computer Science, Stanford University, Stanford, CA, USA}

\icmlcorrespondingauthor{Naoki Murata}{naoki.murata@sony.com}

\icmlkeywords{Diffusion models, Score-based models, Inverse problems, Gibbs sampling}
\vskip 0.3in
]



\printAffiliationsAndNotice{}  

\begin{abstract}
Pre-trained diffusion models have been successfully used as priors in a variety of linear inverse problems, where the goal is to reconstruct a signal from noisy linear measurements. However, existing approaches require knowledge of the linear operator. In this paper, we propose GibbsDDRM, an extension of Denoising Diffusion Restoration Models (DDRM) to a blind setting in which the linear measurement operator is unknown. GibbsDDRM constructs a joint distribution of the data, measurements, and linear operator by using a pre-trained diffusion model for the data prior, and it solves the problem by posterior sampling with an efficient variant of a Gibbs sampler. The proposed method is problem-agnostic, meaning that a pre-trained diffusion model can be applied to various inverse problems without fine-tuning. In experiments, it achieved high performance on both blind image deblurring and vocal dereverberation tasks, despite the use of simple generic priors for the underlying linear operators.
\end{abstract}

\section{Introduction}

\begin{figure}[htb]
\begin{center}
\centerline{\includegraphics[width=0.9\columnwidth]{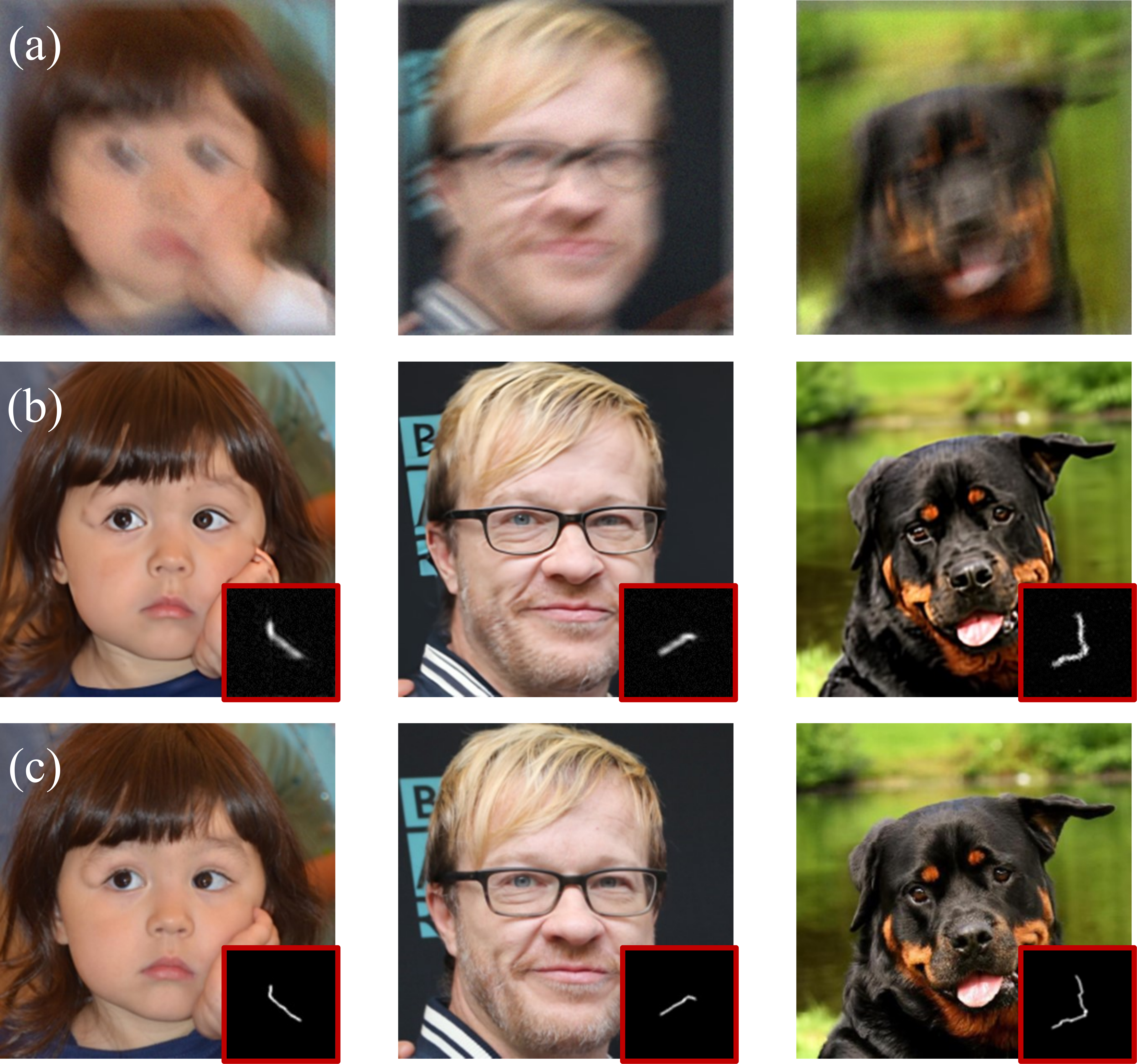}}
\caption{Blind image deblurring results obtained by GibbsDDRM: (a) measurement, (b) restored clean images with blur kernels (bottom right insets), and (c) ground truth images and blur kernels.
}
\label{fig:res_deblur}
\end{center}
\vskip -0.3in
\end{figure}
Inverse problems are frequently encountered in various science and engineering fields such as image processing, acoustic signal processing, and medical imaging. In an inverse problem, the goal is to restore a clean data signal from measurements generated by some forward (measurement) process. In image processing, problems such as deblurring~\cite{zhu2018image, kupyn2019deblurgan, tu2022maxim}, inpainting~\cite{yeh2017semantic}, and colorization~\cite{larsson2016learning} are naturally formulated as inverse problems. In audio signal processing, problems such as dereverberation~\cite{nakatani2010speech, saito2023unsupervised} and band extension~\cite{larsen2005audio} are also classic inverse problems. In medical imaging, many problems such as computed tomography (CT)~\cite{zhu2018image, song2021solving} also rely on inverse problem solving. 

In general, inverse problems are ill-posed because the information in the original data is lost through the measurement process (e.g., because of noise)\camred{;}  
the incorporation of prior knowledge about the original data is thus critical. In the past, assumptions such as sparsity~\cite{candes2008introduction}, low rank~\cite{fazel2008compressed}, and total variation~\cite{candes2006robust} were made for the data distribution to narrow the set of plausible candidate solutions. A more recent trend has been to solve inverse problems by using richer deep generative models~\cite{rick2017one, anirudh2018unsupervised, kadkhodaie2020solving, whang2021solving} trained with a large amount of data as priors. In particular, the evolution of methods related to diffusion models~\cite{kawar2021snips, kawar2022denoising, chung2023diffusion, chung2023parallel} has been significant, and many such methods are problem-agnostic, meaning that they do not require retraining of the generative model used for inference on each task (i.e., each inverse problem).


 Existing approaches typically assume that the measurement process is known. However, many settings are blind, \camred{meaning that} the measurement process itself is (partially) unknown. This is known as a blind setting and includes problems such as blind image deblurring~\cite{pan2016blind} and audio dereverberation ~\cite{nakatani2010speech}. For example, in a blind image deblurring problem, the original image has to be restored from the convolution process where the blur kernel is unknown.
 To address this additional uncertainty, priors are introduced on both the data and the parameters of the linear operator involved~\cite{chan1998total, krishnan2009fast, xu2013unnatural}. BlindDPS~\cite{chung2023parallel} is a method that uses a pre-trained diffusion model for both data and parameters. However, while it can leverage widely available pre-trained diffusion models for signals such as images and audio, it requires training a diffusion model for the parameters of the linear operators of interest, severely restricting its applicability in practice.

To overcome this limitation, we propose GibbsDDRM, which does not require a data-driven prior model of the measurement process. This method is an extension of Denoising Diffusion Restoration Models (DDRM)~\cite{kawar2022denoising} -- a method designed for non-blind linear inverse problems -- to the blind linear setting. 
It constructs a joint distribution of the data, the measurements, and the linear operator's parameters by using a pre-trained diffusion model for the data and a generic prior for the measurement parameters.
Then, it performs approximate sampling from the corresponding posterior distribution of the data and parameters conditioned on the measurements. Here, we adopt a partially collapsed Gibbs sampler (PCGS)~\cite{van2008partially} to enable efficient sampling from the posterior distribution.
PCGS allows us to replace an intractable conditional distribution in the na\"ive Gibbs sampler with a more tractable one without changing the stationary distribution. PCGS alternately samples the data or latent variables and the linear operator's parameters, and the generative model's representational power is exploited while sampling the parameters of the linear operator. This allows our method to accurately estimate both data and the parameters despite using a simple prior for the parameters.

We conducted experiments on the tasks of blind image deblurring in the image processing domain and vocal dereverberation in the acoustic signal processing domain. The results  confirm that high performance can be achieved on both tasks without strong assumptions on the prior for the linear operator's parameters. In the blind image deblurring task, GibbsDDRM demonstrates exceptional quantitative performance in terms of both image quality and faithfulness. It outperforms competing methods and BlindDPS by a large margin in LPIPS, which measures the perceptual similarity of images. The results also show that a faithful image can be restored even with large measurement noise. (see Figure~\ref{fig:res_deblur} for restored images and estimated blur kernels.)  In vocal dereverberation, GibbsDDRM outperforms alternative methods in terms of the quality of the processed vocal, the proximity of the signals, and the degree of reverberation removal.  




\section{Background}
\label{sec:background}
\paragraph{Blind linear inverse problems.}
Blind linear inverse problems involve the estimation of both unknown clean data and the parameters of a linear operator from noisy measurements. This type of problem can be formulated as a linear system of equations of the following form:
\begin{align}
    \mathbf{y} = \mathbf{H}_{\bm{\varphi}}\mathbf{x}_{0} + \mathbf{z},
    \label{eq:mes_model}
\end{align}
where $\mathbf{y}\in \mathbb{R}^{d_{\mathbf{y}}}$ is a vector of measurements, $\mathbf{H}_{\bm{\varphi}}\in\mathbb{R}^{d_{\mathbf{y}}\times d_{\mathbf{x}_{0}}}$ is a linear operator parameterized by $\bm{\varphi}\in\mathbb{R}^{d_{\varphi}}$, \camred{and} $\mathbf{x}_{0}\in\mathbb{R}^{d_{\mathbf{x}_{0}}}$ is the unknown original clean data to be estimated. $\mathbf{z} \sim \mathcal{N}(\mathbf{0}, \sigma_{\mathbf{y}}^{2}\mathbf{I})$ is a Gaussian measurement noise with known covariance $\sigma_{\mathbf{y}}^{2}\mathbf{I}$, where $\mathbf{I}$ is the identity matrix. For notational convenience, we index the clean data $\mathbf{x}_{0}$ with ``$0$" to distinguish it from latent variables of the diffusion model that are defined later. The aim here is to find estimates of both $\mathbf{x}_{0}$ and $\bm{\varphi}$ that fit the given noisy measurements $\mathbf{y}$. 
The problem is ill-posed without any additional assumptions. To obtain a solution, it is assumed that $\mathbf{x}_{0}$ is drawn from a generative model $p_{\theta}(\mathbf{x}_{0})$ (close to the true data distribution), 
and that \camred{the parameters $\bm{\varphi}$ are} drawn from a known prior $p(\bm{\varphi})$ independently \camred{of} the data.
In the Bayesian framework, the optimal solution is to sample from the posterior $p(\mathbf{x}_{0}, \bm{\varphi}|\mathbf{y})$.




\paragraph{Denoising Diffusion Probabilistic Models.}
Denoising Diffusion Probabilistic Models~\cite{sohl2015deep, ho2020denoising, song2019generative, song2021scorebased, Lai2022ImprovingSD},
or diffusion models for short, are generative models with a Markov chain $\mathbf{x}_{T}\rightarrow \dots \rightarrow \mathbf{x}_{t} \rightarrow \dots \rightarrow \mathbf{x}_{0}$ represented by the following joint distribution:
\begin{align}
    p_{\theta}(\mathbf{x}_{0:T}) = p_{\theta}^{(T)}(\mathbf{x}_{T})\prod_{t=0}^{T-1}p_{\theta}^{(t)}(\mathbf{x}_{t}|\mathbf{x}_{t+1}),
    \label{eq:p_ddpm}
\end{align}
where the model's output is $\mathbf{x}_{0}$. To train a diffusion model, a fixed variational inference distribution is introduced:
\begin{align}
    q(\mathbf{x}_{1:T}|\mathbf{x}_{0}) = q^{(T)}(\mathbf{x}_{T}|\mathbf{x}_{0})\prod_{t=1}^{T-1}q^{(t)}(\mathbf{x}_{t}|\mathbf{x}_{t+1}, \mathbf{x}_{0}),
\end{align}
which gives the evidence lower bound (ELBO) on the maximum likelihood objective. With Gaussian parameterization for $p_{\theta}$ and $q$, the ELBO objective is reduced to the following denoising autoencoder objective:
\begin{align}
    \sum_{t=1}^{T}\gamma_{t}\mathbb{E}_{(\mathbf{x}_{0}, \mathbf{x}_{t})\sim p_{\text{data}}(\mathbf{x}_{0})q(\mathbf{x}_{t}|\mathbf{x}_{0})}\left[\|\mathbf{x}_{0}-f_{\theta}^{(t)}(\mathbf{x}_{t})\|_{2}^{2}\right].
\end{align}
Here, $f_{\theta}^{(t)}$ is a $\theta$-parameterized neural network that estimates noiseless data $\mathbf{x}_{0}$ from noisy $\mathbf{x}_{t}$ and characterizes $p_{\theta}$; \nm{$\mathbf{x}_{\theta, t}$ denotes the estimate of noise-less data by $f_{\theta}^{(t)}$;} $\gamma_{t}$ are positive weighting coefficients determined by $q$.

\paragraph{Denoising Diffusion Restoration Models.}
Denoising Diffusion Restoration Models (DDRM)~\cite{kawar2022denoising}
is a method that uses a pre-trained diffusion model as a prior for data in a non-blind linear inverse problem. It is defined as a Markov chain $\mathbf{x}_{T}\rightarrow \mathbf{x}_{T-1} \rightarrow \dots \rightarrow \mathbf{x}_{1} \rightarrow \mathbf{x}_{0}$ (where $\mathbf{x}_{t} \in\mathbb{R}^{d_{\mathbf{x}_{0}}}$) conditioned on the measurements $\mathbf{y}$:
\begin{align}
    p(\mathbf{x}_{0:T}|\mathbf{y}) = p_{\theta}^{(T)}(\mathbf{x}_{T}|\mathbf{y})\prod_{t=0}^{T-1}p_{\theta}^{(t)}(\mathbf{x}_{t}|\mathbf{x}_{t+1}, \mathbf{y}),
\end{align}
where $\mathbf{x}_{0}$ is the model's output. The conditionals in DDRM are defined in terms of the denoising function $f_{\theta}^{(t)}$ of a pre-trained diffusion model; intriguingly, the objective derived \camred{using} the ELBO coincides with that of the unconditional diffusion model, except for a constant factor. This means that the unconditionally pre-trained diffusion model can be used during inference without finetuning.
The core idea of DDRM is to use the singular value decomposition (SVD) of a linear operator $\mathbf{H}$ to transform both the unknown input $\mathbf{x}_{0}$ and the observed output $\mathbf{y}$, \camred{potentially} corrupted by noise, to a shared spectral space. In this space, DDRM executes denoising on dimensions for which information from $\mathbf{y}$ is available (i.e., when the singular values are non-zero). When such information is not available (i.e., when the singular values are zero or the noise in the dimension is large), DDRM performs imputation while explicitly considering the measurement noise. 

\paragraph{Partially collapsed Gibbs sampler.}
A Gibbs sampler is a simple, widely used Markov chain Monte Carlo method for sampling from the joint distribution of a set of variables~\cite{casella1992explaining}. The procedure entails iterative sampling from the fully conditional distributions of each variable, given the current values of the other variables. A blocked Gibbs sampler~\cite{liu1994covariance} is a variant in which, instead of sampling each variable individually, variables in a group or a ``block" of variables are sampled simultaneously while conditioned on all the other variables. This approach is effective when the variables within a block are highly correlated, and it can improve the sampler's convergence speed.

A partially collapsed Gibbs sampler (PCGS)~\cite{van2008partially, kail2012blind} is a generalization of a blocked Gibbs sampler that effectively explores the probability space through three basic operations in the sampling procedure: \textit{marginalization}, \textit{permutation}, and \textit{trimming}, which are described in detail in~\cite{van2008partially} and Appendix~\ref{append:proofs}. In short, the removal of certain variables among the conditional variables does not alter the Gibbs sampler's stationary distribution, as long as these variables are not included among the conditional variables until the next time they are sampled. Hence, we can achieve efficient sampling when the distributions obtained after trimming are tractable.


\section{GibbsDDRM: Partially Collapsed Gibbs Sampler with DDRM}

\subsection{Target joint distribution for blind linear inverse problems} 
\begin{figure}[tb]
\begin{center}
\centerline{\includegraphics[width=0.8\columnwidth]{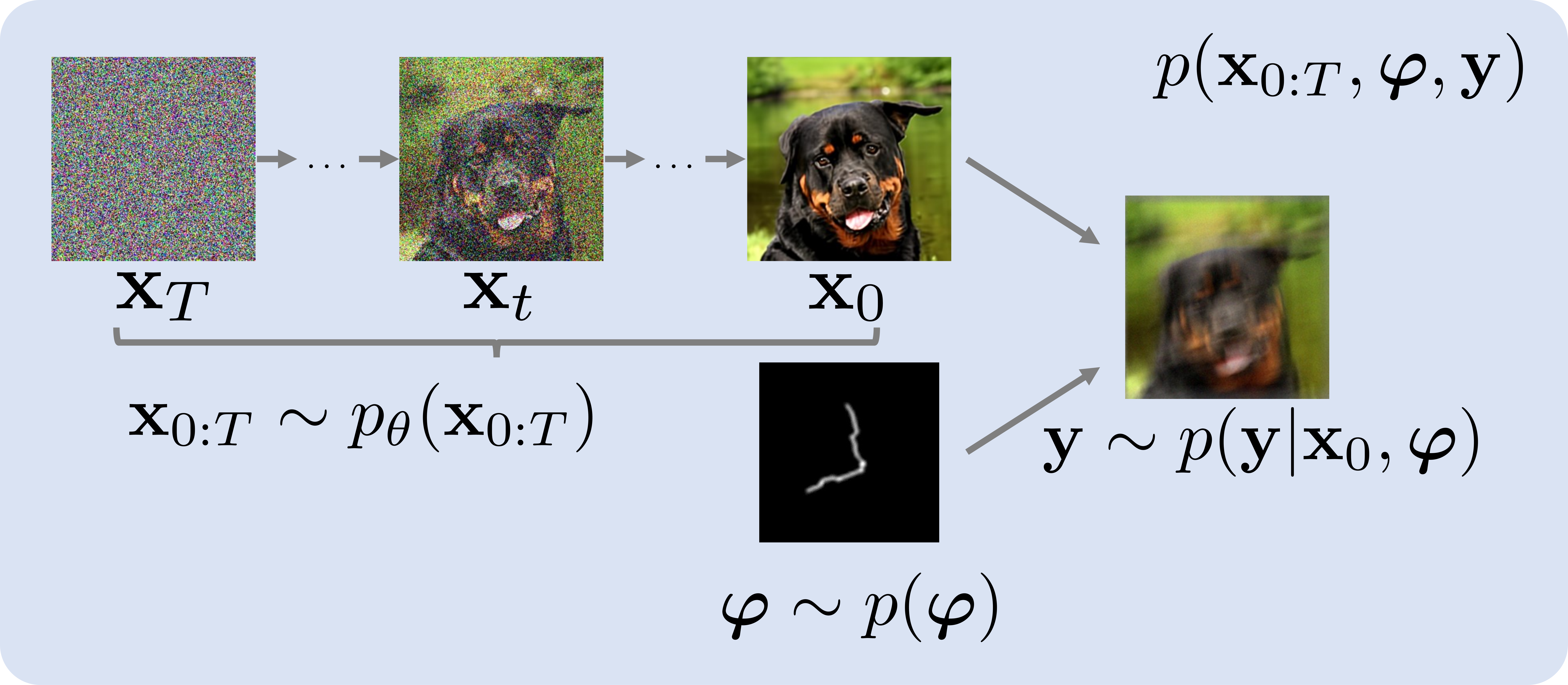}}
\vskip -0.1in
\caption{Graphical model for the joint distribution in Eq.~\eqref{eq:joint_for_all}.}
\label{fig:graphical_model}
\end{center}
\vspace{-24pt}
\end{figure}

In this paper, we seek to solve blind linear inverse problems by sampling from the posterior of the joint distribution of the data and the linear operator's parameters, given the measurements. The joint distribution of the data $\mathbf{x}_{0}$, parameters $\bm{\varphi}$, and measurements $\mathbf{y}$ is defined as follows:
\begin{align}
    p(\mathbf{x}_{0}, \mathbf{y}, \bm{\varphi}) = p_{\theta}(\mathbf{x}_{0})p(\bm{\varphi})\mathcal{N}(\mathbf{y}|\mathbf{H}_{\varphi}\mathbf{x}_{0}, \sigma_{\mathbf{y}}^{2}\mathbf{I}),
    \label{eq:joint_three}
\end{align}
where $p_{\theta}(\mathbf{x}_{0})$ and $p(\bm{\varphi})$ are the known prior distributions for the data and the parameters, respectively. The Gaussian distribution $\mathcal{N}(\mathbf{y}|\mathbf{H}_{\varphi}\mathbf{x}_{0}, \sigma_{\mathbf{y}}^{2}\mathbf{I})$ comes from the measurement model given in Eq.~\eqref{eq:mes_model}. The aim is to sample from the joint posterior distribution $p(\mathbf{x}_{0}, \bm{\varphi}|\mathbf{y})$.  
Using a pre-trained generative model as a prior $p_{\theta}(\mathbf{x}_{0})$ can drastically improve the solutions in inverse problems; however, inference can be challenging. Even in the non-blind setting where $\bm{\varphi}$ is known, sampling from the posterior is intractable and requires approximations like in DDRM~\cite{kawar2022denoising}.



Here we model the data distribution using a pre-trained diffusion model as in Eq.~\eqref{eq:p_ddpm}. This leads to the following joint distribution over the data, its latent variables, and the parameters, as shown in Figure~\ref{fig:graphical_model},
\begin{align}
    &\ p(\mathbf{x}_{0:T}, \bm{\varphi}, \mathbf{y}) \nonumber \\ 
    &= p_{\theta}^{(T)}(\mathbf{x}_{T})\prod_{t=0}^{T-1}p_{\theta}^{(t)}(\mathbf{x}_{t}|\mathbf{x}_{t+1})p(\bm{\varphi})\mathcal{N}(\mathbf{y}|\mathbf{H}_{\bm{\varphi}}\mathbf{x}_{0}, \sigma_{\mathbf{y}}^{2}\mathbf{I}).
    \label{eq:joint_for_all}
\end{align}
Note that sampling from the posterior distribution $p(\mathbf{x}_{0:T}|\bm{\varphi}, \mathbf{y})$ under a fixed $\bm{\varphi}$ corresponds to the objective of DDRM. In addition, we also assume that the parameters' prior $p(\bm{\varphi})$ is a generic 
 and simple prior, such as a sparsity prior.

\subsection{Partially Collapsed Gibbs Sampler for the joint distribution}
To sample from the joint posterior in Eq.~\eqref{eq:joint_for_all}, we could attempt to sample from the joint posterior distribution that includes the latent variables of the diffusion model. However, it is still not feasible to run a na\"ive Gibbs sampler for the posterior $p(\mathbf{x}_{0:T}, \bm{\varphi}|\mathbf{y})$, as it would require a conditional distribution for every individual variable, conditioned on all the other variables. For instance, the conditional distribution $p(\mathbf{x}_{t}|\mathbf{x}_{0:t-1}, \mathbf{x}_{t+1:T}, \bm{\varphi}, \mathbf{y})$ for the joint distribution defined in Eq.~\eqref{eq:joint_for_all} is not obvious.

A possible strategy is to use a blocked Gibbs sampler~\cite{liu1994covariance} with the variables divided into two groups, $\mathbf{x}_{0:T}$ and $\bm{\varphi}$, and sampled alternately.
In more detail, after initializing $\bm{\varphi}$, the sampling procedure of DDRM is performed keeping  $\bm{\varphi}$ fixed to obtain an estimate of the clean data  $\mathbf{x}_0$. Then, $\bm{\varphi}$ is sampled such that it is consistent with the estimated data $\mathbf{x}_0$ and measurements $\mathbf{y}$. By repeating these operations, we can sample $\mathbf{x}_0$ and $\bm{\varphi}$ from the joint posterior.
However, this approach may be inefficient because of the small number of updates made to $\bm{\varphi}$: the entire sampling of $\mathbf{x}_{0:T}$ must be performed for a step of sampling $\bm{\varphi}$, which results in slow convergence. 

Hence, we adopt a partially collapsed Gibbs sampler (PCGS)~\cite{van2008partially} for the joint posterior. This strategy's main advantage is that we can still use a similar sampling method defined by the original DDRM. This enables simultaneous sampling of the latent variables $\mathbf{x}_{1:T}$ and the linear operator's parameters $\bm{\varphi}$ within a cycle of DDRM sampling, thus improving the convergence speed.

In a na\"ive Gibbs sampler, the order of sampling variables is arbitrary. In a PCGS, however, the sampling order must be carefully chosen to facilitate the trimming operation, which removes conditional variables from the conditional distribution. Specifically, once a variable has been marginalized and removed from the conditional set, it should not be added back until the next time it is sampled. We show a simple example of a PCGS in Appendix~\ref{append:proofs}.
 Figure~\ref{fig:fig_gibbs} shows the sampling order of the proposed PCGS. After sampling $\mathbf{x}_{T}$, the following operations are performed in descending order of $t$, until $t=0$: for each $t$, $\mathbf{x}_{t}$ is sampled once, and then $\bm{\varphi}$ and $\mathbf{x}_{t}$ are alternately sampled $M_t$ times. One set of these operations constitutes a single cycle of the PCGS, and the operations are repeated for $N$ cycles.


\begin{figure}[tb]
\begin{center}
\centerline{\includegraphics[width=\columnwidth]{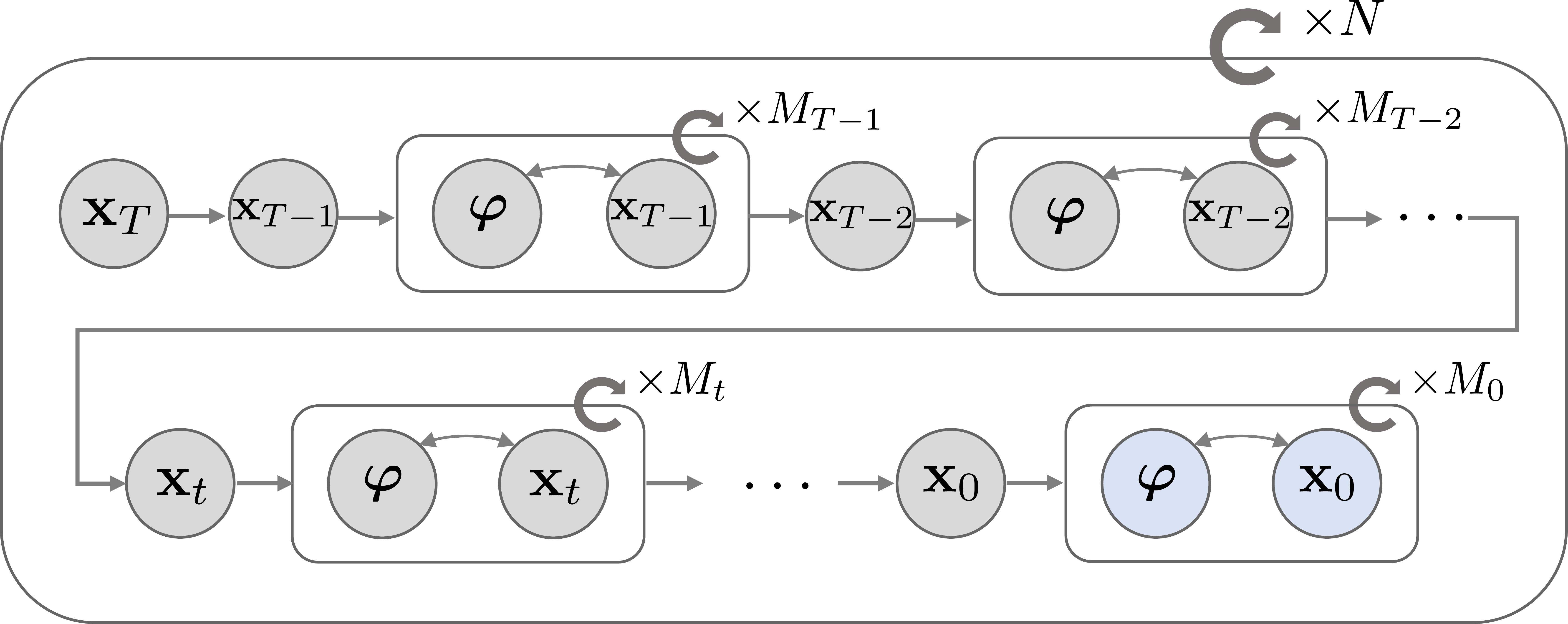}}
\caption{Sampling order of variables in the proposed PCGS, whose output entails the final sample of data $\mathbf{x}_{0}$ and parameters $\bm{\varphi}$.}
\label{fig:fig_gibbs}
\end{center}
\vskip -0.2in
\end{figure}

The proposed PCGS is defined in Algorithm~\ref{alg:example}. The following proposition ensures that it samples from the true posterior distribution.
\begin{proposition}
\label{prop:st_dist_PCGS}
The PCGS defined in Algorithm~\ref{alg:example} has the true posterior distribution $p(\mathbf{x}_{0:T}, \bm{\varphi}|\mathbf{y})$ as its stationary distribution if the approximations to the conditional distributions are exact.
\se{what does it mean they are compatible? does it mean the approximations used are exact?}


\end{proposition}
We give the proof in Appendix~\ref{append:proofs}.

\begin{algorithm}[htb]
   \caption{Proposed PCGS for the posterior in Eq.~\eqref{eq:joint_for_all}}
   \label{alg:example}
\begin{algorithmic}
   \STATE {\bfseries Input:} Measurement $\mathbf{y}$, initial values $\bm{\varphi}^{(0, 0)}$.
   \md{\STATE {\bfseries Output:} Restored data $\mathbf{x}_{0}^{(N, M_0)}$, linear operator's parameters $\bm{\varphi}^{(N, K)}$.}
   \STATE $K \leftarrow 0$      // $K$ counts the number of updates for $\bm{\varphi}$ in a cycle.
   \FOR{$n=1$ {\bfseries to} $N$}
   \STATE $\bm{\varphi}^{(n, 0)} \leftarrow \bm{\varphi}^{(n-1, K)}$, $K \leftarrow 0$
    \STATE Sample $\mathbf{x}_{T}^{(n, 0)}\sim \md{p(\mathbf{x}_{T}|\bm{\varphi}^{(n, K)}, \mathbf{y})}$ 
    \STATE \md{// $\uparrow$ approximated by $p_{\theta}(\mathbf{x}_{T}|\bm{\varphi}, \mathbf{y})$.}
    \FOR{$t=T-1$ {\bfseries to} 0}
    \STATE $\chi_{t} \leftarrow \{\mathbf{x}_{t+1}^{(n, M_{t+1})}, \mathbf{x}_{t+2}^{(n, M_{t+2})}, \cdots, \mathbf{x}_{T}^{(n, 0)}\}$
    \STATE Sample $\mathbf{x}_{t}^{(n, 0)}\sim \md{p(\mathbf{x}_{t}|\bm{\varphi}^{(n, K)}, \chi_{t},  \mathbf{y})}$
    \STATE \md{ // $\uparrow$ approximated by $p_{\theta}(\mathbf{x}_{t}|\mathbf{x}_{t+1}, \bm{\varphi}, \mathbf{y})$. }
        \FOR{$m=1$ {\bfseries to} $M_{t}$}
            \STATE Sample $\bm{\varphi}^{(n, K+1)}\sim \md{p(\bm{\varphi}|\mathbf{x}_{t}^{(n, m-1)}, \chi_{t},  \mathbf{y})}$
            \STATE \md{// $\uparrow$ Langevin sampling with the approximated score $\nabla_{\bm{\varphi}}\log p(\mathbf{y}|\mathbf{x}_{\theta, t}, \bm{\varphi})$.}
            \STATE $K \leftarrow K+1$
            \STATE Sample $\mathbf{x}_{t}^{(n, m)}\sim \md{p(\mathbf{x}_{t}|\bm{\varphi}^{(n, K)}, \chi_{t}, \mathbf{y})}$
            \STATE \md{// $\uparrow$ approximated by $p_{\theta}(\mathbf{x}_{t}|\mathbf{x}_{t+1}, \bm{\varphi}, \mathbf{y})$.}
        \ENDFOR
    \ENDFOR
   \ENDFOR
\end{algorithmic}
\end{algorithm}

Proposition~\ref{prop:st_dist_PCGS} states that it is possible to sample reasonable data and parameters by executing the PCGS defined in Algorithm~\ref{alg:example}, but the conditional distributions \camred{the PCGS includes} are intractable. Hence, we replace each conditional distribution with approximations \camred{from which we can efficiently sample}. In the following paragraphs, we provide the details of the sampling procedures at each step.

\paragraph{Sampling of $\mathbf{x}_{T}$.}
The sampling of $\mathbf{x}_T$ is performed with the distribution \md{$p(\mathbf{x}_{T}|\bm{\varphi}, \mathbf{y})$}, which is obtained by trimming $\mathbf{x}_{0:T-1}$. Because this conditional distribution is intractable, as discussed above, we use modified DDRM to approximate the conditional distribution.

 Here, in order to introduce the modified DDRM, we use SVD of the linear operator $\mathbf{H}_{\bm{\varphi}}$ and its spectral space, similarly to previous studies~\cite{kawar2021snips, kawar2022denoising}. The SVD is given as $\mathbf{H}_{\bm{\varphi}} = \mathbf{U}_{\bm{\varphi}}\mathbf{\Sigma}_{\bm{\varphi}}\mathbf{V}_{\bm{\varphi}}^{\sT}$, where $\mathbf{U}_{\bm{\varphi}}\in\mathbb{R}^{d_{\mathbf{y}}\times d_{\mathbf{y}}}$ and $\mathbf{V}_{\bm{\varphi}}\in\mathbb{R}^{d_{\mathbf{x}_{0}}\times d_{\mathbf{x}_{0}}}$ are orthogonal matrices, and $\mathbf{\Sigma_{\bm{\varphi}}}\in\mathbb{R}^{d_{\mathbf{y}}\times d_{\mathbf{x}_{0}}}$ is a rectangular diagonal matrix. Here we assume $d_{\mathbf{y}} \leq d_{\mathbf{x}_{0}}$, but our method would work for $d_{\mathbf{y}} > d_{\mathbf{x}_{0}}$. The diagonal elements of $\mathbf{\Sigma}_{\bm{\varphi}}$ are the singular values of $\mathbf{H}_{\bm{\varphi}}$ in descending order, denoted $s_{1, \bm{\varphi}}, s_{2, \bm{\varphi}}, \cdots, s_{d_{\mathbf{y}}, \bm{\varphi}}$. Hereafter, we omit the subscript $\bm{\varphi}$ from the singular values for notational simplicity. The values in the spectral space are represented as follows: $\overline{\mathbf{x}}_{t}^{(i)}$ is the $i$-th element of $\overline{\mathbf{x}}_{t} = \mathbf{V}_{\bm{\varphi}}^{\sT}\mathbf{x}_{t}$, and $\overline{\mathbf{y}}^{(i)}$ is the $i$-th element of $\overline{\mathbf{y}} = \mathbf{\Sigma}_{\bm{\varphi}}^{\dagger}\mathbf{U}_{\bm{\varphi}}^{\sT}\mathbf{y}$, where $\mathbf{A}^{\dagger}$ is the Moore-Penrose pseudo-inverse of a matrix $\mathbf{A}$. Note that the spectral space also depends on the parameters $\bm{\varphi}$, which is unknown in our blind setting, unlike in DDRM. 
 Our modified DDRM update for sampling $\mathbf{x}_{T}$ is defined as follows:

\begin{align}
&p_{\theta}^{(T)}\left(\overline{\mathbf{x}}_{T}^{(i)} \mid \mathbf{y}, \bm{\varphi} \right)= \nonumber \\ & \begin{cases}\mathcal{N}\left(\overline{\mathbf{y}}^{(i)}, \sigma_{T}^{2}-\sigma_{\mathbf{y}}^{2} / s_{i}^{2}\right) & \text { if } s_{i}>0 \\ \mathcal{N}\left(0, \sigma_{T}^{2}\right) & \text { if } s_{i}=0\end{cases}\label{eq:gibbsDDRM_pT},
\end{align}
where the only difference from the original DDRM is that the parameters $\bm{\varphi}$ are treated as random variables.

\paragraph{Sampling of $\mathbf{x}_{t}$.}
The sampling of $\mathbf{x}_t$ ($t < T$) is performed by sampling from  the conditional distribution \md{$p(\mathbf{x}_{t} |\mathbf{x}_{t+1:T}, \bm{\varphi}, \mathbf{y})$}, which trims $\mathbf{x}_{0:t-1}$ if $t > 0$. \md{As in the sampling of $\mathbf{x}_{T}$, we approximate the conditional distribution by modifying DDRM. Denoting the prediction of $\mathbf{x}_{0}$ at every time step $t$ by $\mathbf{x}_{\theta, t}$ which is made by the diffusion model as in Sec.~\ref{sec:background}, modified DDRM is defined as follows:
\begin{align}
&p_{\theta}^{(t)}\left(\overline{\mathbf{x}}_{t}^{(i)} \mid \mathbf{x}_{t+1}, \bm{\varphi}, \mathbf{y}\right)= \nonumber \\
& \begin{cases}\mathcal{N}\left(\overline{\mathbf{x}}_{\theta, t}^{(i)}+\sqrt{1-\eta^{2}} \sigma_{t} \frac{\overline{\mathbf{x}}_{t+1}^{(i)}-\overline{\mathbf{x}}_{\theta, t}^{(i)}}{\sigma_{t+1}}, \eta^{2} \sigma_{t}^{2}\right) & \text { if } s_{i}=0 \\
\mathcal{N}\left(\overline{\mathbf{x}}_{\theta, t}^{(i)}+\sqrt{1-\eta^{2}} \sigma_{t} \frac{\overline{\mathbf{y}}^{(i)}-\overline{\mathbf{x}}_{\theta, t}^{(i)}}{\sigma_{\mathbf{y}} / s_{i}}, \eta^{2} \sigma_{t}^{2}\right) & \text { if } \sigma_{t}<\frac{\sigma_{\mathbf{y}}}{s_{i}} \\
\mathcal{N}\left(\left(1-\eta_{b}\right) \overline{\mathbf{x}}_{\theta, t}^{(i)}+\eta_{b} \overline{\mathbf{y}}^{(i)}, \sigma_{t}^{2}-\frac{\sigma_{\mathbf{y}}^{2}}{s_{i}^{2}} \eta_{b}^{2}\right) & \text { if } \sigma_{t} \geq \frac{\sigma_{\mathbf{y}}}{s_{i}}\end{cases}, \label{eq:gibbsDDRM_pt}
\end{align}
where $0 \leq \eta \leq 1$ and $0 \leq \eta_{b} \leq 1$ are hyperparameters, and $0=\sigma_{0} < \sigma_{1} < \sigma_{2} < \dots < \sigma_{T}$ are noise levels that is the same as that defined with the pre-trained diffusion model. Thus we have the approximation 
\begin{align}
    p(\mathbf{x}_{t} |\mathbf{x}_{t+1:T}, \bm{\varphi}, \mathbf{y}) &\simeq p_{\theta}(\mathbf{x}_{t} |\mathbf{x}_{t+1:T}, \bm{\varphi}, \mathbf{y}) \nonumber \\
    &= p_{\theta}(\mathbf{x}_{t}|\mathbf{x}_{t+1}, \bm{\varphi}, \mathbf{y}),
\end{align}
where the final equation comes from the Markov property of the modified DDRM.
}

\begin{figure*}[htb]
\begin{center}
\centerline{\includegraphics[width=6.1in]{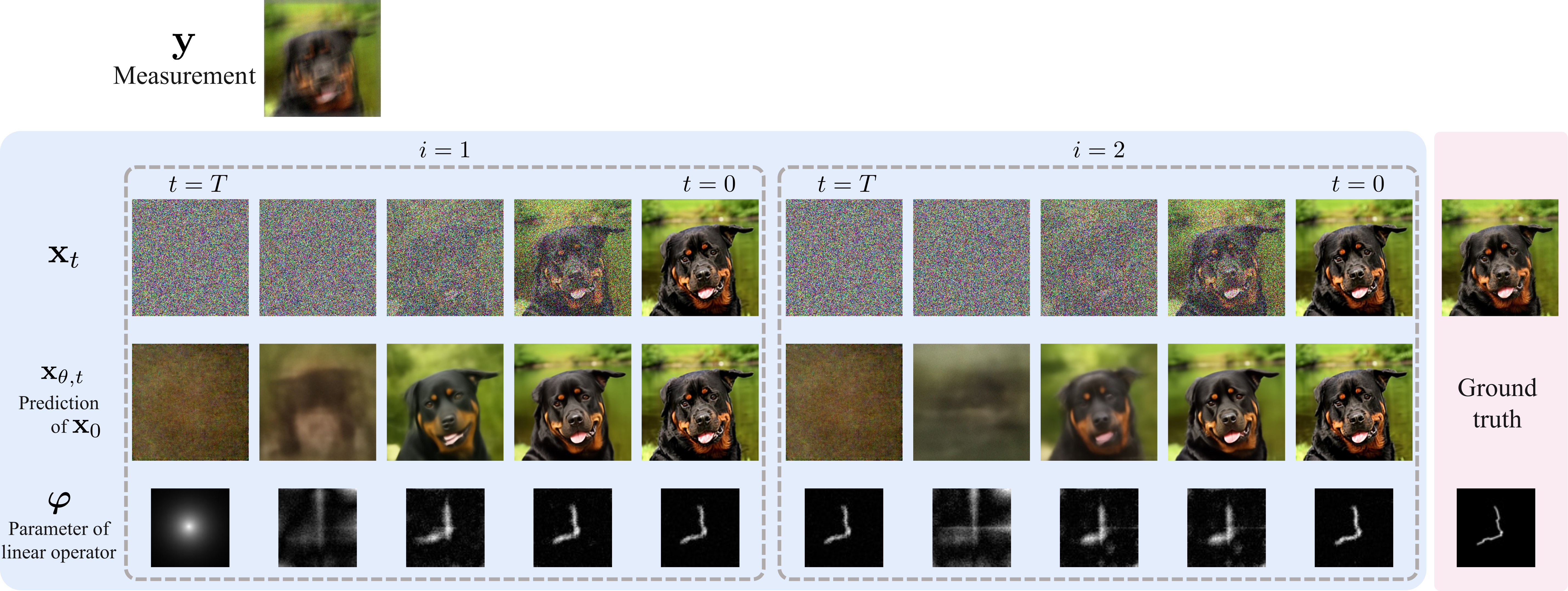}}
\caption{Visualization of GibbsDDRM for the blind image deblurring task on the AFHQ dataset.}
\label{fig:progress_gibbs}
\end{center}
\vspace{-16pt}
\end{figure*}

\paragraph{Sampling of $\bm{\varphi}$.}
\md{At time step $t$, the sampling of the parameters $\bm{\varphi}$ is done by using the conditional distribution $p(\bm{\varphi}|\mathbf{x}_{t:T}, \mathbf{y})$. For the joint distribution defined by Eq.~\eqref{eq:joint_for_all}, the conditional distribution is not easily  obtained because, while $\bm{\varphi}$ and $\mathbf{x}_{t:T}$ are related through $\mathbf{x}_{0}$, the distribution of $\mathbf{x}_{0}$ cannot be evaluated at this point. Hence, we use the approximation in~\cite{chung2023diffusion, chung2023parallel} for the score of the conditional distribution and then perform sampling by Langevin dynamics~\cite{langevin1908theory}, as follows:
\begin{align}
    \bm{\varphi} \leftarrow \bm{\varphi} + (\xi/2)\nabla_{\mathbf{\bm{\varphi}}}\log p(\bm{\varphi}|\mathbf{x}_{t:T}, \mathbf{y}) + \sqrt{\xi}\bm{\epsilon}, 
    \label{eq:phi_langevin}
\end{align}
where $\xi$ is a step size and $\bm{\epsilon}\sim\mathcal{N}(\mathbf{0}, \bm{I})$. By Bayes' rule, the score $\nabla_{\mathbf{\bm{\varphi}}}\log q(\bm{\varphi}|\mathbf{x}_{t:T}, \mathbf{y})$ can be decomposed into two terms:
\begin{align}
    &\nabla_{\mathbf{\bm{\varphi}}}\log p(\bm{\varphi}|\mathbf{x}_{t:T}, \mathbf{y}) =\nonumber \\& \ \ \ \nabla_{\bm{\varphi}}\log p(\mathbf{y}|\mathbf{x}_{t:T}, \bm{\varphi}) + \nabla_{\bm{\varphi}}\log p(\bm{\varphi
}|\mathbf{x}_{t:T}).
\label{eq:decomp_score}
\end{align}
Regarding the first term, we exploit the following theorem.
\begin{theorem}
\label{thm:approx_p_phi}
(modified version of Theorem 1 in \cite{chung2023diffusion}) For the measurement model in Eq.~\eqref{eq:mes_model}, we have
\begin{align}
    p(\mathbf{y}|\mathbf{x}_{t:T}, \bm{\varphi}) \simeq p(\mathbf{y}|\mathbf{x}_{\theta, t}, \bm{\varphi}), 
\end{align}
and the approximation error can be quantified with the Jensen gap~\cite{gao2017bounds}, which is upper bounded by
\begin{align}
    \mathcal{J} \leq \camred{\frac{1}{\sigma_{\mathbf{y}}\left(\sqrt{2\pi \sigma_{\mathbf{y}}^2}\right)^{d_{\mathbf{y}}}}e^{-1/2}}s_{1}m_{1},
\end{align}
where $m_1:=\int\|\mathbf{x}_{0}-\mathbf{x}_{\theta, t}\|p(\mathbf{x}_{0}|\mathbf{x}_{t:T})d{\mathbf{x}_{0}}$, and $s_{1}$ is the largest singular value of $\mathbf{H}_{\bm{\varphi}}$. 
\end{theorem}
By leveraging Theorem~\ref{thm:approx_p_phi}, we obtain the approximate gradient with respect to $\bm{\varphi}$ for the Langevin dynamics:
\begin{align}
    \nabla_{\bm{\varphi}}\log p(\mathbf{y}|\mathbf{x}_{t:T}, \bm{\varphi}) \simeq  \nabla_{\bm{\varphi}}\log p(\mathbf{y}|\mathbf{x}_{\theta, t}, \bm{\varphi}),
\end{align}
and for our measurement model in Eq.~\eqref{eq:mes_model}, the gradient is
\begin{align}
    \nabla_{\bm{\varphi}}\log p(\mathbf{y}|\mathbf{x}_{\theta, t}, \bm{\varphi}) = -\frac{1}{2\sigma_{\mathbf{y}}^2}\nabla_{\bm{\varphi}}\|\mathbf{y} - \mathbf{H}_{\bm{\varphi}}\mathbf{x}_{\theta, t}\|_2^{2},
    \label{eq:score_phi_langevin}
\end{align}
which is tractable in practice.}

\md{As for the second term in Eq.~\eqref{eq:decomp_score}, the conditional variables can be eliminated since $\mathbf{x}_{t:T}$ and $\bm{\varphi}$ are independent from Eq.~\eqref{eq:joint_for_all}. As a result, we can use a simple prior distribution (e.g., a Gaussian prior) for $\bm{\varphi}$ that does not depend on $\mathbf{x}_{t:T}$. }


We now have the conditional score of $\bm{\varphi}$ for the Langevin dynamics as follows:
\begin{align}
    &\nabla_{\mathbf{\bm{\varphi}}}\log p(\bm{\varphi}|\mathbf{x}_{t:T}, \mathbf{y}) \nonumber \\ &\ \simeq -\frac{1}{2\sigma_{\mathbf{y}}^2}\nabla_{\bm{\varphi}}\|\mathbf{y} - \mathbf{H}_{\bm{\varphi}}\mathbf{x}_{\theta, t}\|_2^{2} + \nabla_{\bm{\varphi}}\log p(\bm{\varphi}).
    \label{eq:est_score_phi}
\end{align}
Note that at a particular time step $t$, $\mathbf{x}_{t}$ varies because of the Gibbs sampling, and so does $\mathbf{x}_{\theta, t}$. This iterative process can be viewed as feeding the information from the diffusion model to the parameter estimation. It allows for accurate parameter estimation even with simple priors.

We refer to the proposed PCGS as the Gibbs Denoising Diffusion Restoration Models (GibbsDDRM), and we describe the details of its instantiation for each of our experimental tasks in Appendix~\ref{append:instant_GibbsDDRM}. 

\subsection{Implementation considerations}
\label{ssec:impl_consider}
\paragraph{Initialization of $\bm{\varphi}$.}
In GibbsDDRM, the initialization for $\bm{\varphi}$ is arbitrary. If an existing simple method can be used to obtain an estimate of $\bm{\varphi}$, then we can use that estimate as the initial value. In our experiments, we initialize the blur kernel with a Gaussian blur kernel in the blind image deblurring task. For the vocal dereverberation task, the parameters are initialized with estimates obtained by the weighted prediction error method (WPE)~\cite{nakatani2010speech}, which is an unsupervised method that is not based on machine learning, to accelerate the convergence speed. 
\se{why not sampling from the prior?}

\paragraph{Dependence of number of iterations, $M_{t}$, on time step.}
When $t$ is large, the estimation of $\mathbf{x}_{0}$ ( $= \mathbf{x}_{\theta, t}$) is difficult because of the large amount of noise in $\mathbf{x}_{t}$. This uncertainty can lead to instability in the sampling of $\bm{\varphi}$. The number of sampling steps for $\bm{\varphi}$ can vary across the diffusion time steps and may even be zero. Accordingly, we use a strategy of not updating $\bm{\varphi}$ when $t$ is large.


\section{Experiments}

\begin{table*}[tb]
\centering
\caption{Blind image deblurring results on FFHQ and AFHQ ($256 \times 256$). The blurred images have additive Gaussian noise with $\sigma_{\mathbf{y}} = 0.02$. ($\ast$) The results for BlindDPS~\cite{chung2023parallel}, as reported in the original paper, are also listed, \camred{although} that method uses a pre-trained score function for blur kernels. The results of DDRM~\cite{kawar2022denoising} with the ground truth kernels (i.e., non-blind setting) are also listed. $\textbf{Bold}$: Best. \camred{\uline{underscore}}: second best.}
\vspace{6pt}
\begin{tabular}{lcccccc}
\hline & \multicolumn{3}{c}{\textbf{FFHQ} ($256 \times 256$)} & \multicolumn{3}{c}{\textbf{AFHQ} ($256 \times 256$)} \\ \cline{2-7} 
\textbf{Method}  & FID$\downarrow$       & LPIPS $\downarrow$      & PSNR $\uparrow$      & FID $\downarrow$      & LPIPS$\downarrow$       & PSNR$\uparrow$       \\ \hline
GibbsDDRM (ours)                          &  \uline{38.71}   &  \textbf{0.115}    & \uline{25.80}    &  \uline{48.00}  &  \textbf{0.197}         &  \uline{22.01}  \\ \hline
MPRNet~\cite{zamir2021multi}              &  62.92   &  \uline{0.211}    & \textbf{27.23}    &  50.43  &  \uline{0.278}    & \textbf{27.02}      \\
DeblurGANv2~\cite{kupyn2019deblurgan}     &  141.55  &  0.320    & 19.86    &  156.92  &  0.429    & 17.64      \\
Pan-DCP~\cite{pan2017deblurring}          &  239.69  &  0.653    & 14.20    &  185.40  &  0.632    & 14.48      \\
SelfDeblur~\cite{ren2020neural}           &  283.69  &  0.859    & 10.44    &  250.20  &  0.840    & 10.34      \\ \hline\hline
BlindDPS~\cite{chung2023parallel}$^{\ast}$        &  \textbf{29.49}  &  0.281    & 22.24    &  \textbf{23.89}  &  0.338    & 20.92      \\ \hline
DDRM~\cite{kawar2022denoising} with GT kernel     &  33.97  &  0.062    & 30.64    &  24.60  &  0.078    & 29.37   \\ \hline
\end{tabular}
\label{tbl:res_img_deblur_FFHQ_AFHQ}
\end{table*}

We demonstrate our approach through two tasks: blind image deblurring in the image processing domain and vocal dereverberation in the audio processing domain.

\subsection{Blind image deblurring.}
The aim of blind image deblurring is to restore a clean image from a noisy blurred image without knowledge of the blur kernel. The details of the problem formulation and its instantiation as a linear inverse problem are given in Appendix~\ref{append:instant_GibbsDDRM}. \camred{Our code is available at \url{https://github.com/sony/gibbsddrm}.}

\paragraph{Experimental settings.}
We conduct experiments on the Flickr Face High Quality (FFHQ) $256\times 256$ dataset~\cite{karras2019style} and the Animal Faces-HQ (AFHQ) $256 \times 256$ dataset~\cite{choi2020stargan}. We use a 1000-image validation set for FFHQ and a 500-image test set for the dog class in AFHQ. All images are normalized to the range $[0, 1]$. The blur type used is motion blur, and blur kernels of size $64\times 64$ are generated via code~\footnote{\url{https://github.com/LeviBorodenko/motionblur}}, with an intensity value of $0.5$. We use the pre-trained diffusion models from~\cite{choi2021ilvr}~\footnote{\url{https://github.com/jychoi118/ilvr_adm}} for FFHQ and from~\cite{dhariwal2021diffusion} for AFHQ, without finetuning for this task. Measurements are generated by convolving the blur kernel with a ground truth image and adding Gaussian noise with $\sigma_{\mathbf{y}} = 0.02$. We use $\eta = 0.80$ and $\eta_{b} = 0.90$ for the \nm{proposed method}. The number of steps, $T$, is set to $100$, and $N$ is set to $1$. Following the discussion in Section~\ref{ssec:impl_consider}, $M_{t}$ is set to 0 for $70 \leq t \leq 100$ and to 3 for $t < 70$. The number of iterations and the step size for Langevin dynamics~(Eq.~\eqref{eq:score_phi_langevin}) are set to $500$ and $1.0\times 10^{-11}$, respectively. \camred{The blur kernel is initialized with a Gaussian blur kernel and normalized to have non-negative values and a sum of 1.0, which remains normalized throughout the processing.} We use a Laplace prior for the parameters $\bm{\varphi}$, which has the form $\nabla_{\bm{\varphi}}\log p(\bm{\varphi}) = -\lambda\nabla_{\bm{\varphi}}\|\bm{\varphi}\|_{1}$ The diversity hyperparameter $\lambda$ is set to $10^{3}$. 

\paragraph{Comparison methods.}
We compare GibbsDDRM with several other methods as baselines. These include MPRNet~\cite{zamir2021multi} and DeblurGANv2~\cite{kupyn2019deblurgan} as supervised learning-based baselines, pan-dark channel prior (Pan-DCP)~\cite{pan2017deblurring} as an optimization-based method, and SelfDeblur~\cite{ren2020neural}, which utilizes deep image prior (DIP) for co-estimation of the data and kernel. We also list the results for BlindDPS~\cite{chung2023parallel} as reported in that paper, \camred{although} it uses a prior for the blur kernel that is trained in a supervised manner, thus giving it an unfair advantage.

\paragraph{Evaluation metrics.}
For quantitative comparison of the different methods, the main metrics are the peak signal-to-noise-ratio (PSNR), the Learned Perceptual Image Patch Similarity (LPIPS)~\cite{zhang2018unreasonable}, and the Fr\'{e}chet Inception Distance (FID)~\cite{heusel2017gans}.

\paragraph{Results.}
\begin{figure}[htb]
\begin{center}
\centerline{\includegraphics[width=\columnwidth]{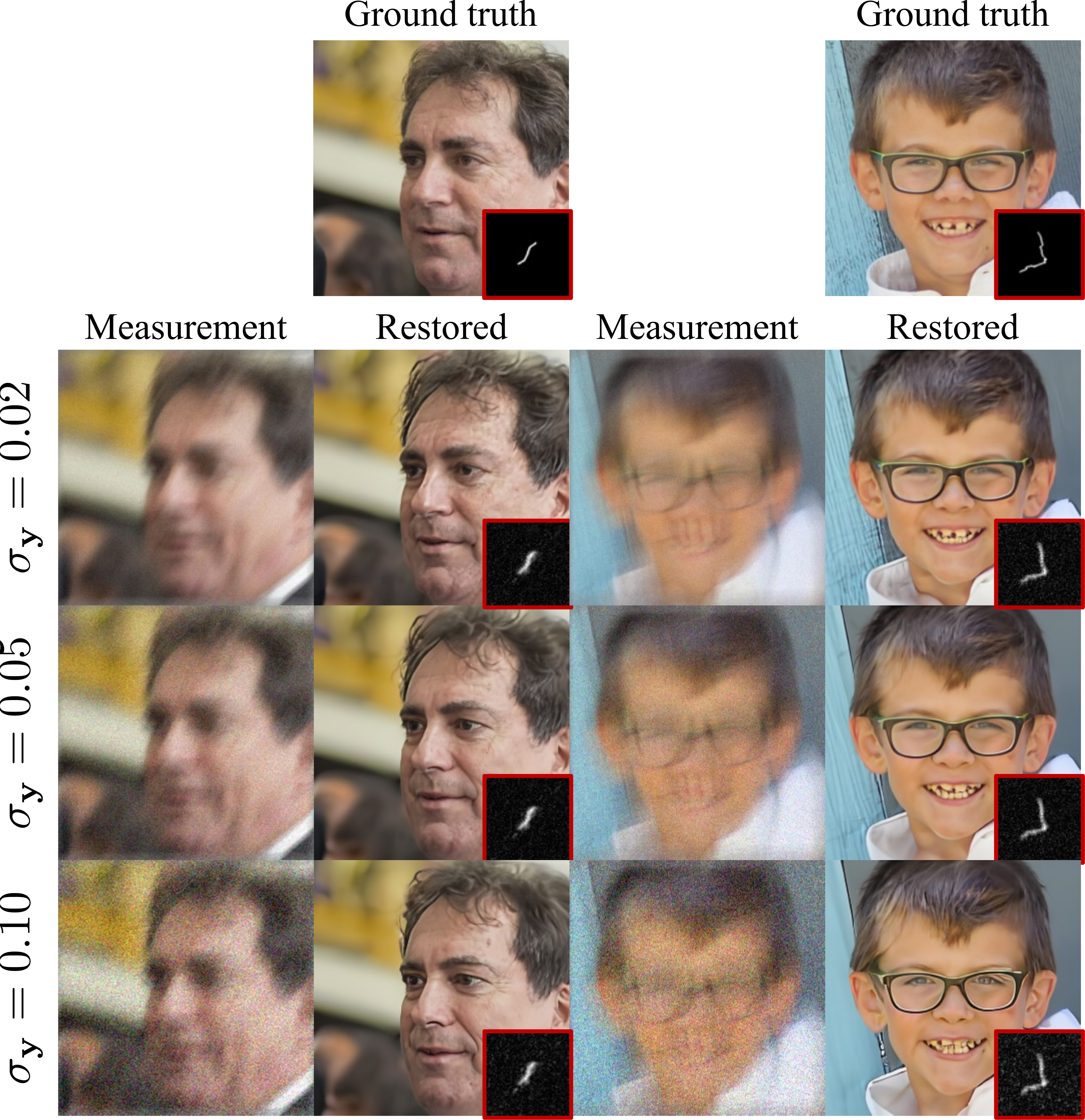}}
\caption{Blurry images and restored images obtained with a restored blur kernel in blind image deblurring under different measurement noise conditions. The top row contains the ground truth images and blur kernels.}
\label{fig:res_deblur_various_noise}
\end{center}
\vspace{-12pt}
\end{figure}

Table~\ref{tbl:res_img_deblur_FFHQ_AFHQ} summarizes the quantitative results of blind image deblurring on FFHQ and AFHQ. \camred{While showing a lower FID score, which measures the quality of generated data, GibbsDDRM outperforms all the other methods in terms of the LPIPS, which measures faithfulness to the original image.} To investigate the performance limit of our method, we also list the results of DDRM with a ground truth kernel.

Figure~\ref{fig:progress_gibbs} visualizes the evolution of the variables for $N=2$. We can see that even in steps where $\mathbf{x}_{t}$ is still quite noisy, the estimated $\mathbf{x}_{\theta, t}$ is close to the ground truth. This leads to an accurate sampling of the blur kernel, which is quite close to the ground truth at $t=0$. 

Figure~\ref{fig:res_deblur_various_noise} shows the restoration results for different measurement noise levels. We can see that even with large noise, a faithful image can be restored via the SVD.

We find that BlindDPS has a lower (better) FID score, but the restored images are relatively far from the original image in terms of the quantitative results. We think that this is because our method uses DDRM, which enables efficient treatment of information obtained from measurements through the SVD, whereas BlindDPS performs more generation than is necessary for noisy observations, which may negatively affect its faithfulness.

Figure~\ref{fig:results_methods} shows the results obtained by our method and the comparison methods. The supervised method MPR achieves the highest PSNR of all the methods, but our method outperforms it in FID and LPIPS.  It is observed that the images obtained by the MPRNet exhibit a certain degree of blurriness when compared to the ground truth images, whereas the images obtained by GibbsDDRM look to be of superior quality in terms of visual perception.

GibbsDDRM takes approximately 56~seconds of computation time per image using one RTX3090 with a batch size of 4. 

\subsection{Vocal dereverberation}
\paragraph{Problem formulation.}
The objective of vocal dereverberation is to restore the original dry vocal from a noisy, reverberant (wet) vocal. Appendix~\ref{append:instant_GibbsDDRM} gives the details of the problem formulation and the specific implementation of the GibbsDDRM that we use for this task.

\paragraph{Experimental settings.}
The proposed method is quantitatively evaluated on wet vocal signals. A pre-trained diffusion model is trained with dry vocal signals from an internal \nm{proprietary} dataset of various genres and singers, with a total duration of 15~hours. A test dataset comprising \ks{1000} wet vocal signals, with a total duration of \ks{around 1.4}~ hours, is prepared by adding artificial reverb to dry vocal signals from the NHSS dataset~\cite{sharma2021nhss}, which contains 100 English pop songs by different singers, with a total duration of 285.24 minutes. Both the training and testing data are monaural recordings sampled at 44.1~kHz. The artificial reverb is added with commercial software by using 10 presets with an RT60 shorter than 2~seconds. The wet vocal signals are prepared by creating $100\times 10$ signals, dividing them into 5-second samples, and randomly selecting \ks{1000} of the resulting signals.




For the GibbsDDRM algorithm, the following parameter values are used: $\eta=0.8$, $\eta_{b}=0.8$, and $\sigma_{y}=1.0 \times 10^{-3}$. We set \ks{$T=50$} for the number of sampling steps and $N=1$. \ks{The parameter $M_{t}$ is set to zero for $40 \leq t \leq 50$, and to \ks{$5$} for $t \leq 40$.} The linear operator's parameters are initialized using results from the WPE algorithm~\cite{nakatani2010speech}, which is an unsupervised method for dereverberation. \ks{The number of iterations and the learning rate for Langevin dynamics (Eq.~\eqref{eq:score_phi_langevin}) are set to $400$ and $1.0 \times 10^{-13}$, respectively. We use a Laplace prior, and the diversity hyperparameter $\lambda$ is set to $2.0$.}
Appendix~\ref{append:exp_settings} gives the details of the network architecture and the dataset.

\paragraph{Comparison methods.}
We evaluate the proposed method against three baselines: Reverb Conversion (RC)~\cite{koo2021reverb}, Music Enhancement (ME)~\cite{kandpal2022music}, and Unsupervised Dereverberation (UD)~\cite{saito2023unsupervised}. RC is a state-of-the-art, end-to-end, DNN-based method that requires pairs of wet and dry vocal signals for dereverberation. It is trained with wet and dry vocal signals that are obtained with different \ks{commercial} reverb plugins from those used for the test dataset. ME is a supervised method based on diffusion models that denoise and dereverb music signals containing vocal signals. It is trained with pairs of 16-kHz reverberant noisy and clean music signals and is evaluated at 16~kHz. UD is a method similar to ours, in that it uses DDRM; however, it differs in how it estimates the linear operator's parameters.

\paragraph{Evaluation metrics.}
For quantitative comparison of the different methods, the metrics are the scale-invariant signal-to-distortion ratio (SI-SDR)~\cite{Roux2019ICASSP} improvement, the Fr\'{e}chet Audio Distance (FAD)~\cite{Kilgour2018Arxiv}, \ks{and the speech-to-reverberation modulation energy ratio (SRMR)~\cite{Santos2014srmr}}. 
Because the FAD uses the pre-trained classification model VGGish~\cite{S_Hershey2017ICASSP}, which is originally trained with $16$~kHz audio samples, we \ks{downsample} all the signals to $16$~kHz to compute FAD.

\paragraph{Results.}
\begin{table}[t]
    \caption{Vocal dereverberation results. $\textbf{Bold}$: Best.}
    \label{tab:vocal_derevereberation}
    \centering
\resizebox{8.5cm}{!}{
    \begin{tabular}{c|c|c|ccc}
    \hline
     \multirow{2}{*}{Method} & \multirow{2}{*}{FAD $\downarrow$} & SI-SDR $\uparrow$ & \multirow{2}{*}{SRMR $\uparrow$} \\
     & & improvement & & \\ \hline
     Wet (unprocessed) & $5.74$ & -- & $7.11$  \\ \hline 
     Reverb Conversion~\cite{koo2021reverb} & $5.69$ & $0.02$ & $7.23$ \\
     Music Enhancement~\cite{kandpal2022music} & $7.51$ & $-23.9$ & $7.92$ \\
     Unsupervised Dereverberation\cite{saito2023unsupervised} & $4.99$ & $0.37$ & $7.94$ \\ \hline
     \textbf{GibbsDDRM} & \bm{$4.21$} & \bm{$0.59$} & \bm{$8.40$} \\ \hline
    \end{tabular}
}
\end{table}

Table~\ref{tab:vocal_derevereberation} lists the scores for each metric. GibbsDDRM outperforms the comparison methods on all metrics. 
In particular, \ks{the result for UD demonstrates that our proposed way of estimating the linear operator's parameters gives better performance than UD's way.} Moreover, \ks{ME doesn't work at all, which may have been because the distribution of its training dataset does not cover that of its test dataset. Indeed, the wet signals for ME's training are created using only simulated natural reverb with some background noise~\cite{kandpal2022music}.}



\section{Conclusion}
We have proposed GibbsDDRM, a method for solving blind linear inverse problems by sampling data and the parameters of a linear operator from a posterior distribution by using a PCGS. The PCGS procedure ensures that the stationary distribution is unchanged from that of the original Gibbs sampler. GibbsDDRM performed well in experiments on blind image deblurring and vocal dereverberation, particularly in terms of preserving the original data, despite its use of a simple prior distribution for the parameters. Additionally, GibbsDDRM has problem-agnostic characteristics, which means that a single pre-trained diffusion model can be used for various tasks. \nm{One limitation of the proposed method is that it is not easily applicable to problems \camred{involving} linear operators for which the SVD is computationally infeasible.}

\section*{Acknowledgements}
We would like to thank Masato Ishii and Stefan Uhlich for their valuable comments during the preparation of this manuscript. Additionally, we thank anonymous reviewers for their insightful suggestions and comments.

{\normalem
\bibliography{str_def_abrv, ref_dgm}

\begin{thebibliography}{63}
\providecommand{\natexlab}[1]{#1}
\providecommand{\url}[1]{\texttt{#1}}
\expandafter\ifx\csname urlstyle\endcsname\relax
  \providecommand{\doi}[1]{doi: #1}\else
  \providecommand{\doi}{doi: \begingroup \urlstyle{rm}\Url}\fi

\bibitem[Anirudh et~al.(2018)Anirudh, Thiagarajan, Kailkhura, and
  Bremer]{anirudh2018unsupervised}
Anirudh, R., Thiagarajan, J.~J., Kailkhura, B., and Bremer, T.
\newblock An unsupervised approach to solving inverse problems using generative
  adversarial networks.
\newblock \emph{{arXiv preprint arXiv}:1805.07281}, 2018.

\bibitem[Cand{\`e}s \& Wakin(2008)Cand{\`e}s and Wakin]{candes2008introduction}
Cand{\`e}s, E.~J. and Wakin, M.~B.
\newblock An introduction to compressive sampling.
\newblock \emph{{IEEE} Signal Process. Mag.}, 25\penalty0 (2):\penalty0 21--30,
  2008.

\bibitem[Cand{\`e}s et~al.(2006)Cand{\`e}s, Romberg, and Tao]{candes2006robust}
Cand{\`e}s, E.~J., Romberg, J., and Tao, T.
\newblock Robust uncertainty principles: Exact signal reconstruction from
  highly incomplete frequency information.
\newblock \emph{{IEEE} Trans. Inf. Theory}, 52\penalty0 (2):\penalty0 489--509,
  2006.

\bibitem[Casella \& George(1992)Casella and George]{casella1992explaining}
Casella, G. and George, E.~I.
\newblock Explaining the {G}ibbs sampler.
\newblock \emph{The American Statistician}, 46\penalty0 (3):\penalty0 167--174,
  1992.

\bibitem[Chan \& Wong(1998)Chan and Wong]{chan1998total}
Chan, T.~F. and Wong, C.-K.
\newblock Total variation blind deconvolution.
\newblock \emph{{IEEE} Trans. Image Process.}, 7\penalty0 (3):\penalty0
  370--375, 1998.

\bibitem[Choi et~al.(2021)Choi, Kim, Jeong, Gwon, and Yoon]{choi2021ilvr}
Choi, J., Kim, S., Jeong, Y., Gwon, Y., and Yoon, S.
\newblock {ILVR}: Conditioning method for denoising diffusion probabilistic
  models.
\newblock In \emph{Proc. IEEE International Conference on Computer Vision
  (ICCV)}, pp.\  14347--14356, 2021.

\bibitem[Choi et~al.(2020{\natexlab{a}})Choi, Kim, Chung, Lee, and
  Jung]{Choi2020ISMIR}
Choi, W., Kim, M., Chung, J., Lee, D., and Jung, S.
\newblock Investigating {U-N}ets with various intermediate blocks for
  spectrogram-based singing voice separation.
\newblock In \emph{Proc. Int. Society for Music Information Retrieval Conf.
  ({ISMIR})}, 2020{\natexlab{a}}.

\bibitem[Choi et~al.(2020{\natexlab{b}})Choi, Uh, Yoo, and Ha]{choi2020stargan}
Choi, Y., Uh, Y., Yoo, J., and Ha, J.-W.
\newblock Star{GAN} v2: Diverse image synthesis for multiple domains.
\newblock In \emph{Proc. IEEE Conference on Computer Vision and Pattern
  Recognition (CVPR)}, pp.\  8188--8197, 2020{\natexlab{b}}.

\bibitem[Chung et~al.(2023{\natexlab{a}})Chung, Kim, Kim, and
  Ye]{chung2023parallel}
Chung, H., Kim, J., Kim, S., and Ye, J.~C.
\newblock Parallel diffusion models of operator and image for blind inverse
  problems.
\newblock In \emph{Proc. IEEE Conference on Computer Vision and Pattern
  Recognition (CVPR)}, 2023{\natexlab{a}}.

\bibitem[Chung et~al.(2023{\natexlab{b}})Chung, Kim, Mccann, Klasky, and
  Ye]{chung2023diffusion}
Chung, H., Kim, J., Mccann, M.~T., Klasky, M.~L., and Ye, J.~C.
\newblock Diffusion posterior sampling for general noisy inverse problems.
\newblock In \emph{Proc. International Conference on Learning Representation
  (ICLR)}, 2023{\natexlab{b}}.

\bibitem[Dhariwal \& Nichol(2021)Dhariwal and Nichol]{dhariwal2021diffusion}
Dhariwal, P. and Nichol, A.
\newblock Diffusion models beat {GAN}s on image synthesis.
\newblock In \emph{Proc. Advances in Neural Information Processing Systems
  (NeurIPS)}, volume~34, pp.\  8780--8794, 2021.

\bibitem[Eaton et~al.(2015)Eaton, Gaubitch, Moore, and Naylor]{Eaton2015ace}
Eaton, J., Gaubitch, N.~D., Moore, A.~H., and Naylor, P.~A.
\newblock The ace challenge — corpus description and performance evaluation.
\newblock In \emph{2015 IEEE Workshop on Applications of Signal Processing to
  Audio and Acoustics (WASPAA)}, pp.\  1--5, 2015.
\newblock \doi{10.1109/WASPAA.2015.7336912}.

\bibitem[Fazel et~al.(2008)Fazel, Candes, Recht, and
  Parrilo]{fazel2008compressed}
Fazel, M., Candes, E., Recht, B., and Parrilo, P.
\newblock Compressed sensing and robust recovery of low rank matrices.
\newblock In \emph{2008 42nd Asilomar Conference on Signals, Systems and
  Computers}, pp.\  1043--1047. IEEE, 2008.

\bibitem[Gao et~al.(2017)Gao, Sitharam, and Roitberg]{gao2017bounds}
Gao, X., Sitharam, M., and Roitberg, A.~E.
\newblock Bounds on the {J}ensen gap, and implications for mean-concentrated
  distributions.
\newblock \emph{{arXiv preprint arXiv}:1712.05267}, 2017.

\bibitem[Hershey et~al.(2017)Hershey, Chaudhuri, Ellis, Gemmeke, Jansen, Moore,
  Plakal, Platt, Saurous, Seybold, Slaney, Weiss, and
  Wilson]{S_Hershey2017ICASSP}
Hershey, S., Chaudhuri, S., Ellis, D. P.~W., Gemmeke, J.~F., Jansen, A., Moore,
  R.~C., Plakal, M., Platt, D., Saurous, R.~A., Seybold, B., Slaney, M., Weiss,
  R.~J., and Wilson, K.
\newblock {CNN} architectures for large-scale audio classification.
\newblock In \emph{Proc. {IEEE} Int. Conf. Acoust., Speech, Signal Process.
  ({ICASSP})}, pp.\  131--135, 2017.

\bibitem[Heusel et~al.(2017)Heusel, Ramsauer, Unterthiner, Nessler, and
  Hochreiter]{heusel2017gans}
Heusel, M., Ramsauer, H., Unterthiner, T., Nessler, B., and Hochreiter, S.
\newblock {GAN}s trained by a two time-scale update rule converge to a local
  {N}ash equilibrium.
\newblock In \emph{Proc. Advances in Neural Information Processing Systems
  (NeurIPS)}, pp.\  6629--6640, 2017.

\bibitem[Ho et~al.(2020)Ho, Jain, and Abbeel]{ho2020denoising}
Ho, J., Jain, A., and Abbeel, P.
\newblock Denoising diffusion probabilistic models.
\newblock \emph{Proc. Advances in Neural Information Processing Systems
  (NeurIPS)}, 33:\penalty0 6840--6851, 2020.

\bibitem[K.~A.~Reddy et~al.(2021)K.~A.~Reddy, Dubey, Koishida, Asokan~Nair,
  Gopal, Cutler, Braun, Gamper, Aichner, and Srinivasan]{reddy2021DNS}
K.~A.~Reddy, C., Dubey, H., Koishida, K., Asokan~Nair, A., Gopal, V., Cutler,
  R., Braun, S., Gamper, H., Aichner, R., and Srinivasan, S.
\newblock Interspeech 2021 deep noise suppression challenge.
\newblock In \emph{Interspeech}, 2021.

\bibitem[Kadkhodaie \& Simoncelli(2020)Kadkhodaie and
  Simoncelli]{kadkhodaie2020solving}
Kadkhodaie, Z. and Simoncelli, E.~P.
\newblock Solving linear inverse problems using the prior implicit in a
  denoiser.
\newblock In \emph{NeurIPS 2020 Workshop on Deep Learning and Inverse
  Problems}, 2020.

\bibitem[Kail et~al.(2012)Kail, Tourneret, Hlawatsch, and
  Dobigeon]{kail2012blind}
Kail, G., Tourneret, J.-Y., Hlawatsch, F., and Dobigeon, N.
\newblock Blind deconvolution of sparse pulse sequences under a minimum
  distance constraint: A partially collapsed {G}ibbs sampler method.
\newblock \emph{{IEEE} Trans. Signal Process.}, 60\penalty0 (6):\penalty0
  2727--2743, 2012.

\bibitem[Kandpal et~al.(2022)Kandpal, Nieto, and Jin]{kandpal2022music}
Kandpal, N., Nieto, O., and Jin, Z.
\newblock Music enhancement via image translation and vocoding.
\newblock In \emph{Proc. {IEEE} Int. Conf. Acoust., Speech, Signal Process.
  ({ICASSP})}, pp.\  3124--3128. IEEE, 2022.

\bibitem[Karras et~al.(2019)Karras, Laine, and Aila]{karras2019style}
Karras, T., Laine, S., and Aila, T.
\newblock A style-based generator architecture for generative adversarial
  networks.
\newblock In \emph{Proc. IEEE Conference on Computer Vision and Pattern
  Recognition (CVPR)}, pp.\  4401--4410, 2019.

\bibitem[Kawar et~al.(2021)Kawar, Vaksman, and Elad]{kawar2021snips}
Kawar, B., Vaksman, G., and Elad, M.
\newblock {SNIPS}: Solving noisy inverse problems stochastically.
\newblock In \emph{Proc. Advances in Neural Information Processing Systems
  (NeurIPS)}, volume~34, pp.\  21757--21769, 2021.

\bibitem[Kawar et~al.(2022)Kawar, Elad, Ermon, and Song]{kawar2022denoising}
Kawar, B., Elad, M., Ermon, S., and Song, J.
\newblock Denoising diffusion restoration models.
\newblock In \emph{Proc. Advances in Neural Information Processing Systems
  (NeurIPS)}, 2022.

\bibitem[Kilgour et~al.(2018)Kilgour, Zuluaga, Roblek, and
  Sharifi]{Kilgour2018Arxiv}
Kilgour, K., Zuluaga, M., Roblek, D., and Sharifi, M.
\newblock {Fr{\'e}chet Audio Distance: A} metric for evaluating music
  enhancement algorithms.
\newblock \emph{{arXiv preprint arXiv}:1812.08466}, 2018.

\bibitem[Koo et~al.(2021)Koo, Paik, and Lee]{koo2021reverb}
Koo, J., Paik, S., and Lee, K.
\newblock Reverb conversion of mixed vocal tracks using an end-to-end
  convolutional deep neural network.
\newblock In \emph{Proc. {IEEE} Int. Conf. Acoust., Speech, Signal Process.
  ({ICASSP})}, pp.\  81--85. IEEE, 2021.

\bibitem[Krishnan \& Fergus(2009)Krishnan and Fergus]{krishnan2009fast}
Krishnan, D. and Fergus, R.
\newblock Fast image deconvolution using hyper-{L}aplacian priors.
\newblock In \emph{Proc. Advances in Neural Information Processing Systems
  (NeurIPS)}, volume~22, 2009.

\bibitem[Kruse et~al.(2017)Kruse, Rother, and Schmidt]{kruse2017learning}
Kruse, J., Rother, C., and Schmidt, U.
\newblock Learning to push the limits of efficient {FFT}-based image
  deconvolution.
\newblock In \emph{Proc. IEEE International Conference on Computer Vision
  (ICCV)}, pp.\  4586--4594, 2017.

\bibitem[Kupyn et~al.(2019)Kupyn, Martyniuk, Wu, and Wang]{kupyn2019deblurgan}
Kupyn, O., Martyniuk, T., Wu, J., and Wang, Z.
\newblock Deblurgan-v2: Deblurring (orders-of-magnitude) faster and better.
\newblock In \emph{Proc. IEEE International Conference on Computer Vision
  (ICCV)}, pp.\  8878--8887, 2019.

\bibitem[Lai et~al.(2022)Lai, Takida, Murata, Uesaka, Mitsufuji, and
  Ermon]{Lai2022ImprovingSD}
Lai, C.-H., Takida, Y., Murata, N., Uesaka, T., Mitsufuji, Y., and Ermon, S.
\newblock Improving score-based diffusion models by enforcing the underlying
  score {F}okker-{P}lanck equation.
\newblock 2022.

\bibitem[Langevin(1908)]{langevin1908theory}
Langevin, P.
\newblock \emph{On the theory of Brownian motion}.
\newblock 1908.

\bibitem[Larsen \& Aarts(2005)Larsen and Aarts]{larsen2005audio}
Larsen, E. and Aarts, R.~M.
\newblock \emph{Audio bandwidth extension: application of psychoacoustics,
  signal processing and loudspeaker design}.
\newblock John Wiley \& Sons, 2005.

\bibitem[Larsson et~al.(2016)Larsson, Maire, and
  Shakhnarovich]{larsson2016learning}
Larsson, G., Maire, M., and Shakhnarovich, G.
\newblock Learning representations for automatic colorization.
\newblock In \emph{European conference on computer vision}, pp.\  577--593.
  Springer, 2016.

\bibitem[Liu et~al.(1994)Liu, Wong, and Kong]{liu1994covariance}
Liu, J.~S., Wong, W.~H., and Kong, A.
\newblock Covariance structure of the {G}ibbs sampler with applications to the
  comparisons of estimators and augmentation schemes.
\newblock \emph{Biometrika}, 81\penalty0 (1):\penalty0 27--40, 1994.

\bibitem[Loshchilov \& Hutter(2019)Loshchilov and Hutter]{Loshchilov2019adamW}
Loshchilov, I. and Hutter, F.
\newblock Decoupled weight decay regularization.
\newblock In \emph{Proc. International Conference on Learning Representation
  (ICLR)}, 2019.

\bibitem[Micikevicius et~al.(2018)Micikevicius, Narang, Alben, Diamos, Elsen,
  Garcia, Ginsburg, Houston, Kuchaiev, Venkatesh, and
  Wu]{Micikevicius2018mixprecison}
Micikevicius, P., Narang, S., Alben, J., Diamos, G., Elsen, E., Garcia, D.,
  Ginsburg, B., Houston, M., Kuchaiev, O., Venkatesh, G., and Wu, H.
\newblock Mixed precision training.
\newblock In \emph{Proc. International Conference on Learning Representation
  (ICLR)}, 2018.

\bibitem[Miyato et~al.(2018)Miyato, Kataoka, Koyama, and
  Yoshida]{miyato2018spectral}
Miyato, T., Kataoka, T., Koyama, M., and Yoshida, Y.
\newblock Spectral normalization for generative adversarial networks.
\newblock In \emph{Proc. International Conference on Learning Representation
  (ICLR)}, 2018.

\bibitem[Nakatani et~al.(2010)Nakatani, Yoshioka, Kinoshita, Miyoshi, and
  Juang]{nakatani2010speech}
Nakatani, T., Yoshioka, T., Kinoshita, K., Miyoshi, M., and Juang, B.-H.
\newblock Speech dereverberation based on variance-normalized delayed linear
  prediction.
\newblock \emph{{IEEE} Trans. Audio, Speech, Lang. Process.}, 18\penalty0
  (7):\penalty0 1717--1731, 2010.

\bibitem[Nichol \& Dhariwal(2021)Nichol and Dhariwal]{Nichol2021improvedDDPM}
Nichol, A. and Dhariwal, P.
\newblock Improved denoising diffusion probabilistic models.
\newblock In \emph{Proc. International Conference on Machine Learning (ICML)},
  pp.\  8162--8171. PMLR, 2021.

\bibitem[Pan et~al.(2016)Pan, Sun, Pfister, and Yang]{pan2016blind}
Pan, J., Sun, D., Pfister, H., and Yang, M.-H.
\newblock Blind image deblurring using dark channel prior.
\newblock In \emph{Proc. IEEE Conference on Computer Vision and Pattern
  Recognition (CVPR)}, pp.\  1628--1636, 2016.

\bibitem[Pan et~al.(2017)Pan, Sun, Pfister, and Yang]{pan2017deblurring}
Pan, J., Sun, D., Pfister, H., and Yang, M.-H.
\newblock Deblurring images via dark channel prior.
\newblock \emph{IEEE transactions on pattern analysis and machine
  intelligence}, 40\penalty0 (10):\penalty0 2315--2328, 2017.

\bibitem[Parmar et~al.(2022)Parmar, Zhang, and Zhu]{parmar2022aliased}
Parmar, G., Zhang, R., and Zhu, J.-Y.
\newblock On aliased resizing and surprising subtleties in gan evaluation.
\newblock In \emph{Proc. IEEE Conference on Computer Vision and Pattern
  Recognition (CVPR)}, pp.\  11410--11420, 2022.

\bibitem[Ren et~al.(2020)Ren, Zhang, Wang, Hu, and Zuo]{ren2020neural}
Ren, D., Zhang, K., Wang, Q., Hu, Q., and Zuo, W.
\newblock Neural blind deconvolution using deep priors.
\newblock In \emph{Proc. IEEE Conference on Computer Vision and Pattern
  Recognition (CVPR)}, pp.\  3341--3350, 2020.

\bibitem[Rick~Chang et~al.(2017)Rick~Chang, Li, Poczos, Vijaya~Kumar, and
  Sankaranarayanan]{rick2017one}
Rick~Chang, J., Li, C.-L., Poczos, B., Vijaya~Kumar, B., and Sankaranarayanan,
  A.~C.
\newblock One network to solve them all--solving linear inverse problems using
  deep projection models.
\newblock In \emph{Proc. IEEE International Conference on Computer Vision
  (ICCV)}, pp.\  5888--5897, 2017.

\bibitem[Ronneberger et~al.(2015)Ronneberger, Fischer, and
  Brox]{Ronneberger2015ICM}
Ronneberger, O., Fischer, P., and Brox, T.
\newblock {U-net: C}onvolutional networks for biomedical image segmentation.
\newblock In \emph{Proceedings of the International Conference on Medical image
  computing and computer-assisted intervention}, pp.\  234--241, 2015.

\bibitem[Roux et~al.(2019)Roux, Wisdom, Erdogan, and Hershey]{Roux2019ICASSP}
Roux, J.~L., Wisdom, S., Erdogan, H., and Hershey, J.~R.
\newblock {SDR – H}alf-baked or well done?
\newblock In \emph{Proc. {IEEE} Int. Conf. Acoust., Speech, Signal Process.
  ({ICASSP})}, pp.\  626--630, 2019.
\newblock \doi{10.1109/ICASSP.2019.8683855}.

\bibitem[Saito et~al.(2023)Saito, Murata, Uesaka, Lai, Takida, Fukui, and
  Mitsufuji]{saito2023unsupervised}
Saito, K., Murata, N., Uesaka, T., Lai, C.-H., Takida, Y., Fukui, T., and
  Mitsufuji, Y.
\newblock Unsupervised vocal dereverberation with diffusion-based generative
  models.
\newblock In \emph{Proc. {IEEE} Int. Conf. Acoust., Speech, Signal Process.
  ({ICASSP})}. IEEE, 2023.

\bibitem[Santos et~al.(2014)Santos, Senoussaoui, and Falk]{Santos2014srmr}
Santos, J.~F., Senoussaoui, M., and Falk, T.~H.
\newblock An improved non-intrusive intelligibility metric for noisy and
  reverberant speech.
\newblock In \emph{Proc. Int. Workshop Acoust. Signal Enhancement ({IWAENC})},
  pp.\  55--59, 2014.
\newblock \doi{10.1109/IWAENC.2014.6953337}.

\bibitem[Sedghi et~al.(2019)Sedghi, Gupta, and Long]{sedghi2018singular}
Sedghi, H., Gupta, V., and Long, P.~M.
\newblock The singular values of convolutional layers.
\newblock In \emph{Proc. International Conference on Learning Representation
  (ICLR)}, 2019.

\bibitem[Sharma et~al.(2021)Sharma, Gao, Vijayan, Tian, and Li]{sharma2021nhss}
Sharma, B., Gao, X., Vijayan, K., Tian, X., and Li, H.
\newblock {NHSS: A} speech and singing parallel database.
\newblock \emph{Speech Communication}, 133:\penalty0 9--22, 2021.

\bibitem[Sohl-Dickstein et~al.(2015)Sohl-Dickstein, Weiss, Maheswaranathan, and
  Ganguli]{sohl2015deep}
Sohl-Dickstein, J., Weiss, E., Maheswaranathan, N., and Ganguli, S.
\newblock Deep unsupervised learning using nonequilibrium thermodynamics.
\newblock In \emph{Proc. International Conference on Machine Learning (ICML)},
  pp.\  2256--2265. PMLR, 2015.

\bibitem[Song \& Ermon(2019)Song and Ermon]{song2019generative}
Song, Y. and Ermon, S.
\newblock Generative modeling by estimating gradients of the data distribution.
\newblock \emph{Proc. Advances in Neural Information Processing Systems
  (NeurIPS)}, 32:\penalty0 11895--11907, 2019.

\bibitem[Song \& Ermon(2020)Song and Ermon]{Song2020improvedSGM}
Song, Y. and Ermon, S.
\newblock Improved techniques for training score-based generative models.
\newblock In \emph{Proc. Advances in Neural Information Processing Systems
  (NeurIPS)}, volume~33, pp.\  12438--12448, 2020.

\bibitem[Song et~al.(2021{\natexlab{a}})Song, Shen, Xing, and
  Ermon]{song2021solving}
Song, Y., Shen, L., Xing, L., and Ermon, S.
\newblock Solving inverse problems in medical imaging with score-based
  generative models.
\newblock In \emph{NeurIPS 2021 Workshop on Deep Learning and Inverse
  Problems}, 2021{\natexlab{a}}.

\bibitem[Song et~al.(2021{\natexlab{b}})Song, Sohl-Dickstein, Kingma, Kumar,
  Ermon, and Poole]{song2021scorebased}
Song, Y., Sohl-Dickstein, J., Kingma, D.~P., Kumar, A., Ermon, S., and Poole,
  B.
\newblock Score-based generative modeling through stochastic differential
  equations.
\newblock In \emph{Proc. International Conference on Learning Representation
  (ICLR)}, 2021{\natexlab{b}}.

\bibitem[Tu et~al.(2022)Tu, Talebi, Zhang, Yang, Milanfar, Bovik, and
  Li]{tu2022maxim}
Tu, Z., Talebi, H., Zhang, H., Yang, F., Milanfar, P., Bovik, A., and Li, Y.
\newblock {MAXIM}: {M}ulti-axis {MLP} for image processing.
\newblock In \emph{Proc. IEEE Conference on Computer Vision and Pattern
  Recognition (CVPR)}, pp.\  5769--5780, 2022.

\bibitem[Van~Dyk \& Park(2008)Van~Dyk and Park]{van2008partially}
Van~Dyk, D.~A. and Park, T.
\newblock Partially collapsed {G}ibbs samplers: Theory and methods.
\newblock \emph{Journal of the American Statistical Association}, 103\penalty0
  (482):\penalty0 790--796, 2008.

\bibitem[Whang et~al.(2021)Whang, Lei, and Dimakis]{whang2021solving}
Whang, J., Lei, Q., and Dimakis, A.
\newblock Solving inverse problems with a flow-based noise model.
\newblock In \emph{Proc. International Conference on Machine Learning (ICML)},
  pp.\  11146--11157. PMLR, 2021.

\bibitem[Xu et~al.(2013)Xu, Zheng, and Jia]{xu2013unnatural}
Xu, L., Zheng, S., and Jia, J.
\newblock Unnatural l0 sparse representation for natural image deblurring.
\newblock In \emph{Proc. IEEE Conference on Computer Vision and Pattern
  Recognition (CVPR)}, pp.\  1107--1114, 2013.

\bibitem[Yeh et~al.(2017)Yeh, Chen, Yian~Lim, Schwing, Hasegawa-Johnson, and
  Do]{yeh2017semantic}
Yeh, R.~A., Chen, C., Yian~Lim, T., Schwing, A.~G., Hasegawa-Johnson, M., and
  Do, M.~N.
\newblock Semantic image inpainting with deep generative models.
\newblock In \emph{Proc. IEEE Conference on Computer Vision and Pattern
  Recognition (CVPR)}, pp.\  5485--5493, 2017.

\bibitem[Zamir et~al.(2021)Zamir, Arora, Khan, Hayat, Khan, Yang, and
  Shao]{zamir2021multi}
Zamir, S.~W., Arora, A., Khan, S., Hayat, M., Khan, F.~S., Yang, M.-H., and
  Shao, L.
\newblock Multi-stage progressive image restoration.
\newblock In \emph{Proc. IEEE Conference on Computer Vision and Pattern
  Recognition (CVPR)}, pp.\  14821--14831, 2021.

\bibitem[Zhang et~al.(2018)Zhang, Isola, Efros, Shechtman, and
  Wang]{zhang2018unreasonable}
Zhang, R., Isola, P., Efros, A.~A., Shechtman, E., and Wang, O.
\newblock The unreasonable effectiveness of deep features as a perceptual
  metric.
\newblock In \emph{Proc. IEEE Conference on Computer Vision and Pattern
  Recognition (CVPR)}, pp.\  586--595, 2018.

\bibitem[Zhu et~al.(2018)Zhu, Liu, Cauley, Rosen, and Rosen]{zhu2018image}
Zhu, B., Liu, J.~Z., Cauley, S.~F., Rosen, B.~R., and Rosen, M.~S.
\newblock Image reconstruction by domain-transform manifold learning.
\newblock \emph{Nature}, 555\penalty0 (7697):\penalty0 487--492, 2018.

\end{thebibliography}
\bibliographystyle{icml2023}
}

\newpage
\appendix
\onecolumn

\section{Proofs}
\label{append:proofs}
\subsection{Proof of Proposition~\ref{prop:st_dist_PCGS}}
\textbf{Proposition}~\ref{prop:st_dist_PCGS}
\textit{The PCGS defined in Algorithm~\ref{alg:example} has the true posterior distribution $p(\mathbf{x}_{0:T}, \bm{\varphi}|\mathbf{y})$ as its stationary distribution if the approximations to the conditional distributions are exact.
\se{what does it mean they are compatible? does it mean the approximations used are exact?}


}

Before giving the proof, we revisit three basic tools for constructing a partially collapsed Gibbs sampler (PCGS)~\cite{van2008partially}. 




\paragraph{Gibbs sampler.}
Let $\bm{\theta} = (\theta_{1}, \dots, \theta_{J})^{\sT}$ be a vector of $J$ variables, and let $\bm{\theta}_{\widetilde{j}}$ denote $\bm{\theta}$ without the $j$~th element $\theta_{j}$. To obtain samples from $p(\bm{\theta})$, a Gibbs sampler~\cite{casella1992explaining} iteratively generates samples of each $\theta_{j}$ from $p(\theta_{j}|\bm{\theta}_{\widetilde{j}})$ in an arbitrary order. \camred{The generated samples approximate the joint distribution of all variables.}


\paragraph{PCGS.} A PCGS is an extension of the Gibbs sampler that facilitates the following three basic tools (see~\cite{van2008partially} for details). 

\begin{itemize}
\item{\textit{Marginalization.}} Rather than sampling only $\theta_{j}$ in a step, other variables may be sampled with $\theta_{j}$ instead of being conditioned on. This process is called marginalization, and it can improve the convergence rate significantly, especially with a strong correlation between the target variables. Within an entire PCGS iteration, certain parameters can be sampled in more than one step. 
\item{\textit{Trimming.}} If a variable is sampled in several steps and is not used as a condition on these steps, only the value sampled in the last step is relevant because the other values are never used. Such unused variables can thus be removed from the respective sampling distribution. This reduces the complexity of the sampling steps without affecting the convergence behavior.
\item{\textit{Permutation.}} It is reasonable to choose an (arbitrary) sampling order such that trimming can be performed. After trimming, permutations are only allowed if they preserve the justification of the trimming that has already been applied.
\end{itemize}
For example, the following PCGS for sampling $(\mathbf{X}, \mathbf{Y}, \mathbf{Z}, \mathbf{W})$ is a simple PCGS. 
\begin{align}
&\text{Step 1. Sample $\mathbf{Y}$ from }p(\mathbf{Y}, \cancel{\mathbf{W}}|\mathbf{X}, \mathbf{Z}) \nonumber \\
&\text{Step 2. Sample $\mathbf{Z}$ from }p(\mathbf{Z}, \cancel{\mathbf{W}}|\mathbf{X}, \mathbf{Y}) \nonumber \\
&\text{Step 3. Sample $\mathbf{W}$ from }p(\mathbf{W}|\mathbf{X}, \mathbf{Y}, \mathbf{Z}) \nonumber \\
&\text{Step 4. Sample $\mathbf{X}$ from }p(\mathbf{X}|\mathbf{W}, \mathbf{Y}, \mathbf{Z})
\end{align}
Here, the random variable $\mathbf{W}$ is trimmed in steps~1 and~2 because it is sampled in step~3 before being included in the conditional variables. Note that the order of steps 3 and 4 cannot be interchanged. The reason is that the variable $\mathbf{W}$, which is trimmed in steps 1 and 2, would be included among the conditional variables in $p(\mathbf{X}|\mathbf{W}, \mathbf{Y}, \mathbf{Z})$, thus altering the sampler's stationary distribution.
\begin{proof}
To show that the proposed sample is a valid PCGS, we transform a na\"ive Gibbs sampler by applying the above PCGS tools to the proposed PCGS. \camred{First,} we consider the na\"ive Gibbs sampler defined in Algorithm~\ref{alg:sampler1}, which we denote as Sampler~1.

Sampler~1 has a stationary distribution $p(\mathbf{x}_{0:T}, \bm{\varphi}|\mathbf{y})$, since it is a na\"ive \camred{Gibbs} sampler for the joint distribution in Eq.~\eqref{eq:joint_for_all}. In Gibbs sampling, the stationary distribution is unaffected by repeating certain steps and changing the order of the steps. \camred{Therefore,} the sampling scheme depicted in Figure~\ref{fig:fig_gibbs} constructs Sampler~2, which is defined in Algorithm~\ref{alg:sampler2}. 

\camred{Next, Sampler~2 is converted to Sampler~3, which is defined in Algorithm~\ref{alg:sampler3}, by marginalizing the variables $\psi_{t}$. Subsequently, we convert the Sampler~3 to our proposed sampler by employing the PCGS trimming operation and approximating the conditional distributions.} The variables $\psi_{t}$ for $t = 0, 1, \dots, T$ can be trimmed from Sampler~3, as they do not appear in the conditional variables of the conditional distribution \camred{before} they are next sampled. Because the proposed PCGS corresponds to a sampler that omits $\psi_{t}$ from Sampler~3, it has the true posterior distribution $p(\mathbf{x}_{0:T}|\mathbf{y})$ as its stationary distribution. Thus, if the approximations for the conditional distributions are exact, the PCGS has the true posterior distribution as its stationary distribution.\qedhere

\begin{algorithm}[htb]
   \caption{Sampler 1 for the posterior in Eq.~\eqref{eq:joint_for_all}}
   \label{alg:sampler1}
\begin{algorithmic}
   \STATE {\bfseries Input:} Measurement $\mathbf{y}$, initial values $\bm{\varphi}^{(0)}$, $\mathbf{x}_{0:T}^{(0)}$.
   \STATE {\bfseries Output:} Restored data $\mathbf{x}_{0}^{(N)}$, linear operator's parameters $\bm{\varphi}^{(N)}$
   \FOR{$n=1$ {\bfseries to} $N$}
    \STATE Sample $\mathbf{x}_{T}^{(n)}\sim p(\mathbf{x}_{T}|\mathbf{x}_{0:T-1}^{(n-1)}, \bm{\varphi}^{(n-1)}, \mathbf{y})$
    \FOR{$t=T-1$ {\bfseries to} 0}
    \STATE Sample $\mathbf{x}_{t}^{(n)}\sim p(\mathbf{x}_{t}|\mathbf{x}_{0:t-1}^{(n-1)}, \mathbf{x}_{t+1:T}^{(n)}, \bm{\varphi}^{(n-1)}, \mathbf{y})$ 
    \ENDFOR
    \STATE $\bm{\varphi}^{(n)}\sim p(\bm{\varphi}|\mathbf{x}_{0:T}^{(n)}, \mathbf{y})$
   \ENDFOR
\end{algorithmic}
\end{algorithm}

\begin{algorithm}[htb]
   \caption{Sampler 2 for the posterior in Eq.~\eqref{eq:joint_for_all}}
   \label{alg:sampler2}
\begin{algorithmic}
   \STATE {\bfseries Input:} Measurement $\mathbf{y}$, initial values $\bm{\varphi}^{(0, 0)}$,
   $\mathbf{x}_{0:T}^{(0, M_{0})}$
   \STATE {\bfseries Output:} Restored data $\mathbf{x}_{0}^{(N, M_0)}$, parameters of linear operator $\bm{\varphi}^{(N, K)}$
   \STATE $K \leftarrow 0$   \ \ \      // $K$ counts the number of updates for $\bm{\varphi}$ in a cycle.
   \FOR{$n=1$ {\bfseries to} $N$}
   \STATE $\bm{\varphi}^{(n, 0)} \leftarrow \bm{\varphi}^{(n-1, K)}$ 
   \STATE $K \leftarrow 0$
    \STATE $\psi_{T} \leftarrow \{\mathbf{x}_{0}^{(n-1, M_{0})}, \mathbf{x}_{1}^{(n-1, M_{1})}, \dots, \mathbf{x}_{t}^{(n-1, M_t)}, \dots, \mathbf{x}_{T-1}^{(n-1, M_{T-1})}\mathbf\}$
    \STATE Sample $\mathbf{x}_{T}^{(n, 0)}\sim p(\mathbf{x}_{T}|\bm{\varphi}^{(n, K)}, \uwave{\psi_{T}}, \mathbf{y})$
    \FOR{$t=T-1$ {\bfseries to} 0}
    \STATE $\psi_{t} \leftarrow \{\mathbf{x}_{0}^{(n-1, M_{0})}, \mathbf{x}_{1}^{(n-1, M_{1})}, \dots, \mathbf{x}_{t-1}^{(n-1, M_{t-1})}\}$
    \STATE $\chi_{t} \leftarrow \{ \mathbf{x}_{t+1}^{(n, M_{t+1})}, \mathbf{x}_{t+2}^{(n, M_{t+2})}, \cdots, \mathbf{x}_{T}^{(n, 0)}\}$
    \STATE Sample $\mathbf{x}_{t}^{(n, 0)}\sim p(\mathbf{x}_{t}|\bm{\varphi}^{(n, K)}, \uwave{\psi_{t}}, \chi_{t},  \mathbf{y})$
        \FOR{$m=1$ {\bfseries to} $M_{t}$}
            \STATE Sample $\bm{\varphi}^{(n, K+1)}\sim p(\bm{\varphi}|\mathbf{x}_{t}^{(n, m-1)}, \uwave{\psi_{t}}, \chi_{t},  \mathbf{y})$ 
            \STATE $K \leftarrow K+1$
            \STATE Sample $\mathbf{x}_{t}^{(n, m)}\sim p(\mathbf{x}_{t}|\bm{\varphi}^{(n, K)}, \uwave{\psi_{t}}, \chi_{t}, \mathbf{y})$
        \ENDFOR
    \ENDFOR
   \ENDFOR
\end{algorithmic}
\end{algorithm}

\begin{algorithm}[htb]
   \caption{\camred{Sampler 3 for the posterior in Eq.~\eqref{eq:joint_for_all}}}
   \label{alg:sampler3}
\begin{algorithmic}
   \STATE {\bfseries Input:} Measurement $\mathbf{y}$, initial values $\bm{\varphi}^{(0, 0)}$,
   $\mathbf{x}_{0:T}^{(0, M_{0})}$
   \STATE {\bfseries Output:} Restored data $\mathbf{x}_{0}^{(N, M_0)}$, parameters of linear operator $\bm{\varphi}^{(N, K)}$
   \STATE $K \leftarrow 0$   \ \ \      // $K$ counts the number of updates for $\bm{\varphi}$ in a cycle.
   \FOR{$n=1$ {\bfseries to} $N$}
   \STATE $\bm{\varphi}^{(n, 0)} \leftarrow \bm{\varphi}^{(n-1, K)}$ 
   \STATE $K \leftarrow 0$
    \STATE $\psi_{T} \leftarrow \{\mathbf{x}_{0}^{(n-1, M_{0})}, \mathbf{x}_{1}^{(n-1, M_{1})}, \dots, \mathbf{x}_{t}^{(n-1, M_t)}, \dots, \mathbf{x}_{T-1}^{(n-1, M_{T-1})}\mathbf\}$
    \STATE Sample \{$\mathbf{x}_{T}^{(n, 0)}, \uwave{\psi_{T}}\}\sim p(\mathbf{x}_{T}|\bm{\varphi}^{(n, K)},  \mathbf{y})$
    \FOR{$t=T-1$ {\bfseries to} 0}
    \STATE $\psi_{t} \leftarrow \{\mathbf{x}_{0}^{(n-1, M_{0})}, \mathbf{x}_{1}^{(n-1, M_{1})}, \dots, \mathbf{x}_{t-1}^{(n-1, M_{t-1})}\}$
    \STATE $\chi_{t} \leftarrow \{ \mathbf{x}_{t+1}^{(n, M_{t+1})}, \mathbf{x}_{t+2}^{(n, M_{t+2})}, \cdots, \mathbf{x}_{T}^{(n, 0)}\}$
    \STATE Sample $\{\mathbf{x}_{t}^{(n, 0)}, \uwave{\psi_{t}}\}\sim p(\mathbf{x}_{t}|\bm{\varphi}^{(n, K)},  \chi_{t},  \mathbf{y})$
        \FOR{$m=1$ {\bfseries to} $M_{t}$}
            \STATE Sample $\{\bm{\varphi}^{(n, K+1)}, \uwave{\psi_{t}}\}\sim p(\bm{\varphi}|\mathbf{x}_{t}^{(n, m-1)}, \chi_{t},  \mathbf{y})$ 
            \STATE $K \leftarrow K+1$
            \STATE Sample $\{\mathbf{x}_{t}^{(n, m)}, \uwave{\psi_{t}}\}\sim p(\mathbf{x}_{t}|\bm{\varphi}^{(n, K)}, \chi_{t}, \mathbf{y})$
        \ENDFOR
    \ENDFOR
   \ENDFOR
\end{algorithmic}
\end{algorithm}
\end{proof}

\newpage
\subsection{Proof of Theorem~\ref{thm:approx_p_phi}}
We follow the result from~\cite{chung2023diffusion, chung2023parallel}. First, we confirm the following lemmas.
\begin{lemma} \label{lemma:L_univariate} Let $\phi(\cdot)$ be a univariate Gaussian density function with mean $\mu$ and variance $\sigma^{2}$. $\phi(\cdot)$ is $L$-Lipschitz such that $\forall x_1, x_2 \in\mathbb{R}$, 
\begin{align}
    |\phi(x_1) - \phi(x_2)| \leq L |x_1-x_2|,
    \label{eq:lip_uni_G}
\end{align}
where $L = \frac{1}{\sqrt{2\pi} \sigma^{2}}e^{-1/2}$.
\end{lemma}
\begin{proof} 
Since $\phi(\cdot)$ is an everywhere differentiable function and it has the bounded first derivative, we use the mean value theorem to get
\begin{align}
    \forall x_1, x_2 \in \mathbb{R}, |\phi(x_1)-\phi(x_2)|\leq \|\phi'\|_{\infty}|x_1-x_2|.
\end{align}
Since $L$ is the minimal value for Eq.~\eqref{eq:lip_uni_G}, we have that $L\leq \|\phi'\|_{\infty}$. Taking the limit $x_2 \rightarrow x_1$ gives $|\phi'(x)|\leq L$, and thus $\|\phi'\|_{\infty}\leq L$. Hence
\begin{align}
    L = \|\phi'\|_{\infty} = \left\|-\frac{x-\camred{\mu}}{\sigma^2}\phi(x)\right\|_{\infty}.
\end{align}
Since the derivative of $\phi'$ is given as
\begin{align}
    \phi''(x) = -\sigma^{-2}(1-\sigma^{-2}(x-\mu)^{2})\phi(x),
\end{align}
and the maximum is attained when $x = \mu \pm{\sigma}$, we have
\begin{align}
    L = \|\phi'\|_{\infty} = \frac{1}{\sqrt{2\pi}\sigma^2}e^{-1/2}
\end{align}
\end{proof}

\begin{lemma}\label{lemma:L_multivariate}
Let $f(\cdot)$ be an isotropic multivariate Gaussian density function with mean $\bm{\mu}$ and variance $\sigma^{2}\mathbf{I}$. $f(\cdot)$ is $L$-Lipschitz such that $\forall \mathbf{x}_1, \mathbf{x}_2 \in\mathbb{R}^{d}$, 
\begin{align}
    \|f(\mathbf{x}_1) - f(\mathbf{x}_2)\| \leq L \|\mathbf{x}_1-\mathbf{x}_2\|,
\end{align}
where 
\begin{align}
L = \frac{1}{\sigma\left(\sqrt{2\pi \sigma^2}\right)^{d}}e^{-1/2}
\end{align}
\end{lemma}
\begin{proof}
\camred{
We first evaluate the value of $\max _{\mathbf{x}}\|\nabla f(\mathbf{x})\|$, where 
$f(\mathbf{x}) = \prod_{i=1}^{d}\phi(x_{i})$. Without loss of generality, we assume $\bm{\mu}=\mathbf{0}$.
\begin{align}
    \nabla f(\mathbf{x}) &= \left[ \frac{\partial f(\mathbf{x})}{\partial x_{1}}, \dots, \frac{\partial f(\mathbf{x})}{\partial x_{d}}\right]^{\sT}\nonumber \\
    &= \left[\phi'(x_{1})\prod_{i\neq 1}\phi(x_{i}), \dots, \phi'(x_{d})\prod_{i\neq d}\phi(x_{i})\right]^{\sT} \nonumber \\
    &= \left[-\frac{x_{1}}{\sigma^{2}}\prod_{i=1}^{d}\phi(x_{i}), \dots, -\frac{x_{d}}{\sigma^{2}}\prod_{i=1}^{d}\phi(x_{i})\right]^{\sT} \nonumber \\
    &= -\frac{\prod_{i=1}^{d}\phi(x_{i})}{\sigma^{2}}\left[x_{1},\dots, x_{d}\right]^{\sT}.
\end{align}
Therefore, $\max _{\mathbf{x}}\|\nabla f(\mathbf{x})\|$ can be evaluated as follows,
\begin{align}
\|\nabla f(\mathbf{x})\| &= \sqrt{x_{1}^{2}+\dots+x_{d}^{2}}\frac{\prod_{i=1}^{d}\phi(x_{i})}{\sigma^{2}} \nonumber \\
&= \sqrt{x_{1}^{2}+\dots+x_{d}^{2}} \frac{\exp\left(-\frac{x_{1}^{2}+\dots+x_{d}^{2}}{2\sigma^2}\right)}{\sigma^{2}\cdot\left(\sqrt{2\pi \sigma^2}\right)^{d}} \nonumber \\
&= r \frac{\exp(-\frac{r^2}{2\sigma^2})}{\sigma^{2}\cdot\left(\sqrt{2\pi \sigma^2}\right)^{d}}, \ \ r\geq 0 \ \ \left(r=\sqrt{x_{1}^{2}+\dots+x_{d}^{2}}\right) \nonumber \\
&\overset{\mathrm{(a)}}{\leq} \frac{1}{\sigma\left(\sqrt{2\pi \sigma^2}\right)^{d}}e^{-1/2} = C_{\text{multi}},
\end{align}
where the equality holds when $r \ (= \sqrt{x_{1}^{2}+\dots+x_{d}^2}) = \sigma$. (a) is by the result of the lemma~\ref{lemma:L_univariate}. Here, by the mean value theorem, for any $\mathbf{x}_1, \mathbf{x}_2\in\mathbb{R}^{d}$, the following holds: 
\begin{align}
    \|f(\mathbf{x}_1) - f(\mathbf{x}_2)\|\leq C_{\text{multi}}\|\mathbf{x}_1-\mathbf{x}_2\|.
\end{align}
By setting $\mathbf{x}_1=[\sigma, 0, \dots, 0]^{\sT}$ and taking the limit $\mathbf{x}_2\rightarrow\mathbf{x}_1$, the equality holds. Hence, $f(\cdot$) is $L$-Lipschitz with the Lipschitz constant $L=C_{\text{multi}}$.
}
\end{proof}


\begin{lemma}
    \label{lemma:L_linear}
    Let $\mathbf{H}\in\mathbb{R}^{d_{\mathbf{y}}\times d_{\mathbf{x}}}$ be a linear operator. The linear operator is $L$-Lipschitz such that $\forall \mathbf{x}_1, \mathbf{x}_2\in\mathbb{R}^{d_{\mathbf{x}}}$,
    \begin{align}
        \|\mathbf{H}\mathbf{x}_1-\mathbf{H}\mathbf{x}_2\|\leq L\|\mathbf{x}_1-\mathbf{x}_2\|,
    \end{align}
    where $L$ is the largest singular value of $\mathbf{H}$.
\end{lemma}
\camred{This property has been reported in several papers, such as~\cite{miyato2018spectral}.}

\textbf{Theorem}~\ref{thm:approx_p_phi}
\textit{(modified version of Theorem 1 in \cite{chung2023diffusion}) For the measurement model in Eq.~\eqref{eq:mes_model}, we have
\begin{align}
    p(\mathbf{y}|\mathbf{x}_{t:T}, \bm{\varphi}) \simeq p(\mathbf{y}|\mathbf{x}_{\theta, t}, \bm{\varphi}), 
\end{align}
and the approximation error can be quantified with the Jensen gap~\cite{gao2017bounds}, which is upper bounded by
\begin{align}
    \mathcal{J} \leq \camred{\frac{1}{\sigma_{\mathbf{y}}\left(\sqrt{2\pi \sigma_{\mathbf{y}}^2}\right)^{d_{\mathbf{y}}}}e^{-1/2}}s_{1}m_{1},
\end{align}
where $m_1:=\int\|\mathbf{x}_{0}-\mathbf{x}_{\theta, t}\|p(\mathbf{x}_{0}|\mathbf{x}_{t:T})d{\mathbf{x}_{0}}$, and $s_{1}$ is the largest singular value of $\mathbf{H}_{\bm{\varphi}}$. }
    
\begin{proof}
In our case, the Jensen gap~\cite{gao2017bounds} is defined as follows:
\begin{align}
    \mathcal{J} &= \left|\camred{p(\mathbf{y}|\mathbf{x}_{t:T}, \bm{\varphi})} - p(\mathbf{y}|\mathbf{x}_{\theta, t}, \bm{\varphi})\right| \nonumber.
\end{align}
Let $f(\bm{\mu})$ be an isotropic multivariate Gaussian density function with mean $\bm{\mu}$ and variance $\sigma_{\mathbf{y}}^{2}\mathbf{I}$, and thus $p(\mathbf{y}|\mathbf{x}_{0}, \bm{\varphi}) = f(\mathbf{H}_{\bm{\varphi}}\mathbf{x}_{0})$ in our case.
The Jensen gap is evaluated as follows:
\begin{align}
    \mathcal{J} &= \left|\camred{p(\mathbf{y}|\mathbf{x}_{t:T}, \bm{\varphi})} - p(\mathbf{y}|\mathbf{x}_{\theta, t}, \bm{\varphi})\right| \nonumber \\
    &= \left| \int \left(p(\mathbf{y}|\mathbf{x}_{t:T}, \mathbf{x}_{0}, \bm{\varphi}) - p(\mathbf{y}|\mathbf{x}_{\theta, t}, \bm{\varphi})\right)p(\mathbf{x}_{0}|\mathbf{x}_{t:T})d\mathbf{x}_{0} \right| \nonumber \\
    &\overset{\mathrm{(a)}}{=} \left| \int \left(p(\mathbf{y}|\mathbf{x}_{0}, \bm{\varphi}) - p(\mathbf{y}|\mathbf{x}_{\theta, t}, \bm{\varphi})\right)p(\mathbf{x}_{0}|\mathbf{x}_{t})d\mathbf{x}_{0} \right| \nonumber \\
    &= \left| \int \left(f(\mathbf{H}_{\bm{\varphi}}\mathbf{x}_{0}) - f(\mathbf{H}_{\bm{\varphi}}\mathbf{x}_{\theta, t})\right)p(\mathbf{x}_{0}|\mathbf{x}_{t})d\mathbf{x}_{0} \right| \nonumber \\
    &\overset{\mathrm{(b)}}{\leq} \camred{\frac{1}{\sigma_{\mathbf{y}}\left(\sqrt{2\pi \sigma_{\mathbf{y}}^2}\right)^{d_{\mathbf{y}}}}e^{-1/2}}\int \|\mathbf{H}_{\bm{\varphi}}\mathbf{x}_{0}-\mathbf{H}_{\bm{\varphi}}\mathbf{x}_{\theta, t}\|p(\mathbf{x}_{0}|\mathbf{x}_{t})d\mathbf{x}_{0} \nonumber \\
    &\overset{\mathrm{(c)}}{\leq} \camred{\frac{1}{\sigma_{\mathbf{y}}\left(\sqrt{2\pi \sigma_{\mathbf{y}}^2}\right)^{d_{\mathbf{y}}}}e^{-1/2}}s_{1}\int \|\mathbf{x}_{0}-\mathbf{x}_{\theta, t}\|p(\mathbf{x}_{0}|\mathbf{x}_{t})d\mathbf{x}_{0} \nonumber \\
    &=\camred{\frac{1}{\sigma_{\mathbf{y}}\left(\sqrt{2\pi \sigma_{\mathbf{y}}^2}\right)^{d_{\mathbf{y}}}}e^{-1/2}}s_{1}m_{1},
\end{align}
where (a) is by the conditional independence of $\mathbf{y}$ and $\mathbf{x}_{t:T}$ given $\mathbf{x}_{0}$ and the Markov property of $\mathbf{x}_{t:T}$, and (b) and (c) are by the lemmas~\ref{lemma:L_multivariate} and \ref{lemma:L_linear}.
\end{proof}

\section{Instantiation of blind linear inverse problems}
\label{append:instant_GibbsDDRM}
\paragraph{Blind image deblurring.} 
The aim of blind image deblurring is to restore a clean image from a noisy blurred image without knowledge of the blur kernel. The problem is formulated as follows:
\begin{align}
    \mathbf{y} = \mathbf{k}\ast \mathbf{x}_{0} + \mathbf{z},
\end{align}
where $\mathbf{k}$ is the blur kernel, corresponding to the parameters $\bm{\varphi}$ in our setting, and $\ast$ denotes the convolution operator. Although dealing with this problem in our framework requires the SVD of the convolution operator, it can be computed efficiently by using an FFT~\cite{sedghi2018singular, kruse2017learning}. Thus, the SVD enables efficient calculation in the spectral domain. In performing the SVD with an FFT, it is necessary to consider signals in the complex domain; however, the proposed method can be naturally extended to the complex case.

\paragraph{Vocal dereverberation.}
The details of dealing with vocal dereverberation as a linear inverse problem are discussed in~\cite{saito2023unsupervised}. Let $y_{\tau, f}^{\text{wet}}\in\mathbb{C}$ be the wet (reverberant) vocal signals in a short-time Fourier transform (STFT) domain, where $\tau$ and $f$ denote the respective time and frequency indices. We use the following measurement model:
\begin{align}
    y_{\tau, f}^{\text{wet}} = \sum_{l= 0}^{L-1}g_{l, f}^{*}x_{\tau -l, f}^{\text{dry}} + z_{\tau, f}, 
\end{align}
where $x_{\tau, f}^{\text{dry}}\in\mathbb{C}$ and $g_{\tau, f}\in\mathbb{C}$ are the dry vocal signals and the acoustic transfer function between wet and dry signals, respectively. Here, we assume additive noise $z_{\tau, f}\in\mathbb{C}$. $(\cdot)^{*}$ denotes the complex conjugate, and $L$ is the length of reverberation. As with blind image deblurring, the linear operator, in this case, is a convolution operator whose acoustic transfer function is unknown. Thus, the efficient method of performing the SVD by using an FFT is applicable.
    
\section{Details on experimental settings}
\label{append:exp_settings}
\subsection{Blind image deblurring.}
\paragraph{Comparison methods.}
For methods requiring training data, images from the dataset are corrupted with blur kernels that are generated by using the MotionBlur library\footnote{\url{https://github.com/LeviBorodenko/motionblur}} and Gaussian noise with variance $\sigma_{\mathbf{y}} = 0.02$ is added. The blur kernel size is $64\times 64$, and the intensity value is determined for each kernel by uniform sampling from the range [0.4, 0.6].

\textbf{MPRNet~\cite{zamir2021multi}.} We use the official implementation~\footnote{\url{https://github.com/swz30/MPRNet}} for the deblurring task, with the recommended parameters, learning rate decay, and neural network architectures. The model is trained for 100k iterations with a batch size of 4 for both the FFHQ and AFHQ datasets.

\textbf{DeblurGANv2~\cite{kupyn2019deblurgan}.}
We use the official implementation~\footnote{\url{https://github.com/VITA-Group/DeblurGANv2}} while adhering to the default settings for the parameters and network architectures. Specifically, the model is trained by minimizing the sum of the pixel distance loss, WGAN-gp adversarial loss, and perceptual loss with the weight parameters specified in the official implementation. The generator uses Inception-ResNet-v2 as its backbone. For both the FFHQ and AFHQ datasets, the model is trained for 500k iterations with a batch size of 1. The hyperparameters for the loss are set to $\lambda_{\text{pixel}} = 5.0\times 10^{-1}$, $\lambda_{\text{adv}}=6.0\times 10^{-3}$, and $\lambda_{\text{perceptual}}=1.0\times 10^{-2}$.

\textbf{Pan-DCP~\cite{pan2016blind}.}
We use the official implementation~\footnote{\url{https://jspan.github.io/projects/dark-channel-deblur}} with the parameters recommended for facial images. For the hyperparameters, we use $\lambda_{\text{dark}} = 4.0\times 10^{-3}$, and $\lambda_{\text{grad}}=4.0 \times 10^{-3}$. The number of iterations is set to $5$.

\textbf{SelfDeblur~\cite{ren2020neural}.}
We use the official implementation~\footnote{\url{https://github.com/csdwren/SelfDeblur}} with the default settings for YCbCR and a fixed learning rate of 0.01 for 2500 steps. The optimization process involves minimizing the mean squared error (MSE) for the initial 500 steps, followed by a switch to the structural similarity index (SSIM) loss function for the remaining steps.

\camred{
\textbf{Details on evaluation metrics.}
The FID scores reported in the paper are calculated using the cleanfid library~\cite{parmar2022aliased}~\footnote{\url{https://github.com/GaParmar/clean-fid}}. Specifically, for FFHQ, the evaluation is conducted with 1,000 restored images and 70,000 images from the training and validation set. Similarly, for AFHQ, the evaluation is conducted on 500 restored images and 4,739 images from the training set. The limited number of samples used in the evaluation is due to the computational complexity of the proposed method. The BlindDPS paper doesn't provide details on the calculation of FID, so there may be slight differences in the reported values.}

\subsection{Vocal dereverberation.}
The pre-trained diffusion model \ks{for GibbsDDRM is} trained with only dry vocal signals from an internal dataset containing various genres of songs by various singers. The total signal duration is around $15$ hours.
For a test dataset, we use \ks{$1000$} wet vocal signals (\ks{1.4}~hours in total) by adding artificial reverb to dry vocal signals from another dataset, the NHSS dataset~\cite{sharma2021nhss}.
That dataset contains $100$ English pop songs ($20$ unique songs) by different singers, with a total signal duration of $285.24$ minutes. 
Each song for training and testing is sampled at $44.1$~kHz and features monaural recording.
For artificial reverb, we use the presets for vocals in the FabFilter Pro-R plug-in~\footnote{\url{https://www.fabfilter.com/products/pro-r-reverb-plug-in}}, \ks{which is a commercial artificial reverb plug-ins}.
From a total of $19$ kinds of vocal reverb presets, we use all the presets whose RT60 is shorter than 2~seconds (10 in total). 
We \ks{prepare wet test dataset} by creating $100 \times 10$ signals, dividing them into $5$-second samples, and randomly selecting \ks{$1000$} of the resulting signals.

The implementation of our method and the network architecture of the pre-trained diffusion model are mostly based on the code provided by the authors of the DDRM paper~\footnote{\url{https://github.com/bahjat-kawar/ddrm}}.
We slightly \ks{modify} certain parts as follows. 
We \ks{convert} each audio input to a complex-valued STFT representation by using a window size of $1024$, a hop size of $256$, and a Hann window.
Further, to follow the original input configuration, we cut the direct-current components of the input signals and input them as $2$-channeled $512 \times 512$ image data. The first and second channels correspond to the respective real and imaginary parts.
We \ks{modify} the original U-Net~\cite{Ronneberger2015ICM} architecture of the pre-trained model used on DDRM by adding a time-distributed, fully connected \ks{(TFC)} layer~\cite{Choi2020ISMIR} to the last layer of every residual block \ks{expecting the TFC layers to capture the harmonic structure of music signals efficiently.}

For the training, we reduce the diffusion model's size by having fewer trainable parameters ($31.3$~M), and the training took less than three days with an NVIDIA A100 GPU.
\ks{The hyperparameters for the training of the diffusion model are in Table~\ref{tab:hpara_dereverb}.
We also incorporate an adaptive group normalization~\cite{dhariwal2021diffusion} into each residual block.
We train the model using AdamW~\cite{Loshchilov2019adamW} with $\beta_{1}=0.9$ and $\beta_{2}=0.999$ in $16$-bit precision~\cite{Micikevicius2018mixprecison}. We use an exponential moving average over model parameters with a rate of $0.9999$~\cite{Song2020improvedSGM}.}
     
\begin{table}[h]
    \caption{\ks{Hyperparameters for training diffusion model on dry vocal signals}. We follow the same notations defined in~\cite{dhariwal2021diffusion}}
        \label{tab:hpara_dereverb}
        \centering
\resizebox{9.0cm}{!}{
    \begin{tabular}{lc}
    \hline 
     Diffusion steps & $4000$ \\
     Noise schedule & cosine~\cite{Nichol2021improvedDDPM} \\
     Model size & $31.3$~M \\  
     Channels & $64$  \\
     Depth & $2$ \\  
     Channels multiple & $1, 1, 2, 2, 4, 4$ \\
     Heads & $2$ \\
     Attention resolution & $32, 16$ \\
     BigGAN up/downsample & \checkmark \\
     Dropout & $0.0$ \\
     Batch size & $6$ \\
     Iterations & $370$K\\
     Learning rate & $1.0\times 10^{-4}$ \\
     \hline
    \end{tabular}
}
\end{table}
\ks{For initialization of the linear operator, we used the WPE with the parameters $L=150$, $D=4$, and one iteration. GibbsDDRM takes 36~seconds to restore 1~second vocal signals, whereas UD takes 6~seconds.}

\paragraph{Comparison methods.}

\noindent \textbf{Reverb conversion}: A state-of-the-art end-to-end DNN-based method for vocal dereverberation. We use the original code and the pre-trained model\footnote{The original code and the pre-trained model are shared by Junghyun Koo from the Department of Intelligence and Information at Seoul National University. Mr. Koo also assisted with the discussion of the RC results of our experiment.}, which is trained with the pairs of $44.1$~kHz wet and dry vocal signals. 
Note that the wet signals are reverbed with the artificial reverb for vocal taken from the different commercial reverb plug-ins \ks{(e.g., \footnote{\url{https://valhalladsp.com/shop/reverb/valhalla-room/}}\footnote{\url{https://valhalladsp.com/shop/reverb/valhalla-vintage-verb/}})} from those of our test dataset~\cite{koo2021reverb}.
We input pairs of wet and dry signals since this method needs them for dereverberation.

\noindent \textbf{Music enhancement}: A supervised method to denoise and dereverb music signals based on diffusion models~\cite{kandpal2022music}. 
We use both the original code and the pre-trained model specified in the paper. 
Since ME is trained with pairs of $16$~kHz reverberant noisy and clean music signals containing vocal signals, we evaluate this method at $16$~kHz for all the objective metrics.
\ks{Note that the wet signals of the training dataset are created using room impulse responses from the DNS Challenge dataset~\cite{reddy2021DNS}, which may have some different characteristics from artificial reverb for vocal signals, and adding the background noise from the ACE Challenge dataset~\cite{Eaton2015ace}.}

\noindent \textbf{UnsupervisedDereverb}: An unsupervised method for vocal dereverberation~\cite{saito2023unsupervised}. 
\ks{This method is similar to our GibbsDDRM, which utilizes DDRM. However, it differs in how it estimates the linear operator's parameter.
We use the same pre-trained diffusion model as GibbsDDRM.
We set $L=150, D=4$, the number of iterations of WPE to one, $\eta=0.8, \eta_{b} = 0.8$, $\sigma_{y} = 1.0 \times 10^{-3}$, with the number of sampling steps $T$ set to $50$.
The number of iterations, the learning rate, and the regularization parameter for refinement of the linear operator are set to $10000$, $1.0 \times 10^{-6}$, and $1.0$, respectively.}

\section{Additional Results.}
\subsection{Blind image deblurring.}
\paragraph{Qualitative comparison.}
\begin{figure}[htb]
\centering
\centerline{\includegraphics[width=6.5in]{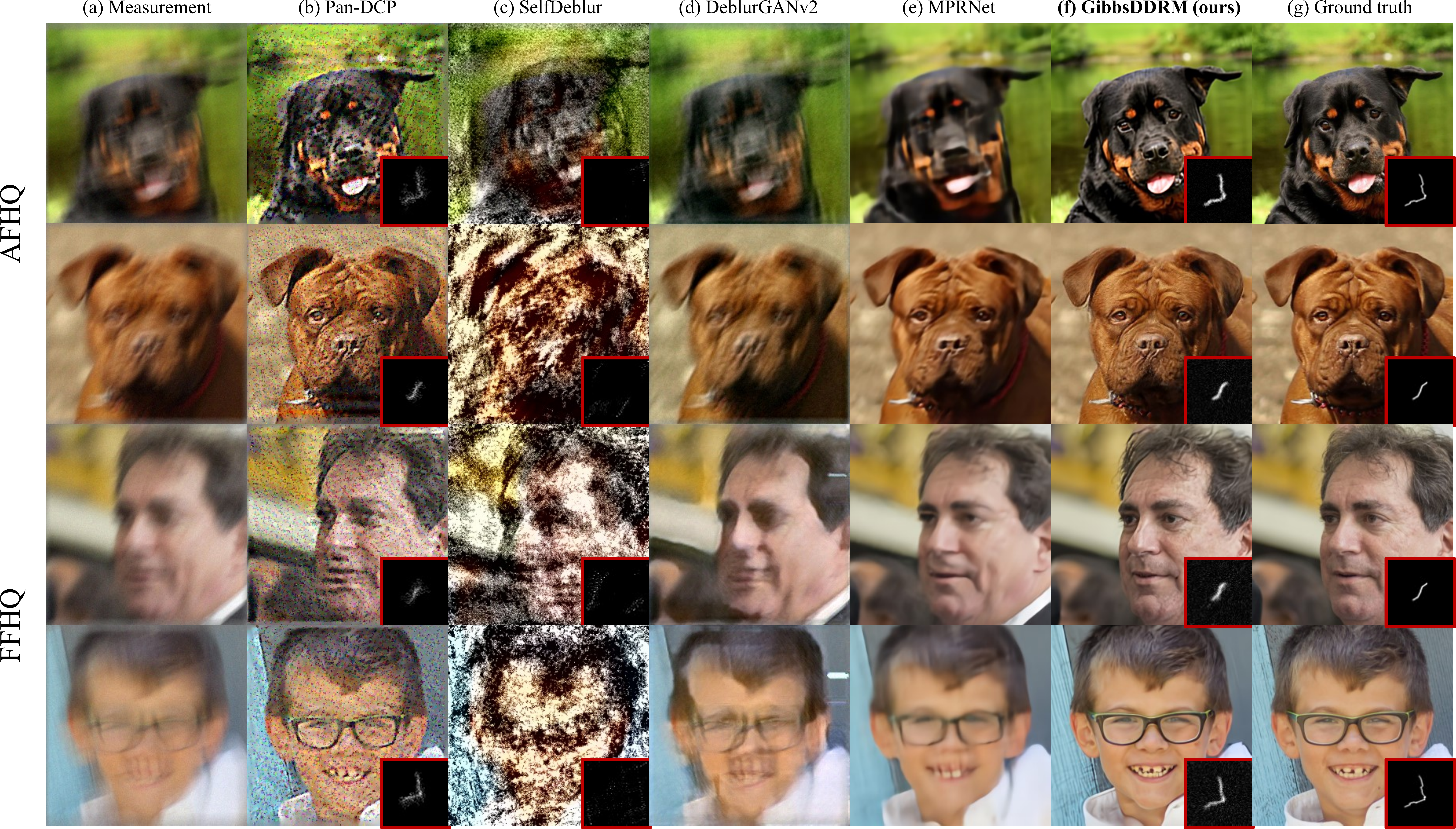}}
\caption{Blind image deblurring results on the FFHQ and AFHQ datasets: (a)~measurements, (b)~Pan-DCP~\cite{pan2016blind}, (c)~SelfDeblur~\cite{ren2020neural}, (d)~DeblurGANv2~\cite{kupyn2019deblurgan}, (e)~MPRNet~\cite{zamir2021multi}, (f)~GibbsDDRM (ours), and (g) ground truth. The kernels are also shown for methods that estimate them.}
\label{fig:results_methods}
\end{figure}
We show the results of our method and comparison methods in Figure.~\ref{fig:results_methods}. The images estimated by GibbsDDRM appear perceptually similar to the ground truth images, but the images estimated by MPRNet have better quality in terms of PSNR. However, the images estimated by MPRNet lack definition compared to the ground truth images. GibbsDDRM utilizes a generative model to generate components lost during the measurement process by considering the spectral space of the linear operator, which is one of the reasons why MPRNet outperforms GibbsDDRM in terms of PSNR. In addition, it is important to note that MPRNet is specifically trained on the corruption caused by motion blur.

In our experiments, other comparison methods, except for MPRNet, do not perform well in restoring the images with a high degree of accuracy. This is consistent with the results reported in~\cite{chung2023parallel}. In the motion blur corruption process used in this study, the blur kernel is relatively large to the image size, and there is also measurement noise, making it challenging to estimate a stable solution in such situations.

\paragraph{Relationship between hyperparameters $\eta$ and $\eta_{b}$ and each evaluation metric.}
\camred{
We show the relationship between the hyperparameters $\eta$ and $\eta_{b}$ and each evaluation metric on the FFHQ dataset in Table~\ref{tab:relationship_etas}. Note that although there is a small difference, the parameters that best achieve LPIPS differ from those that best achieve PSNR. The parameters for $T$, $M_{t}$, and Langevin dynamics are set to be the same as those described in the paper.}

\begin{figure}[htb]
\centering
\begin{subfigure}[b]{0.45\textwidth}
\includegraphics[width=\textwidth]{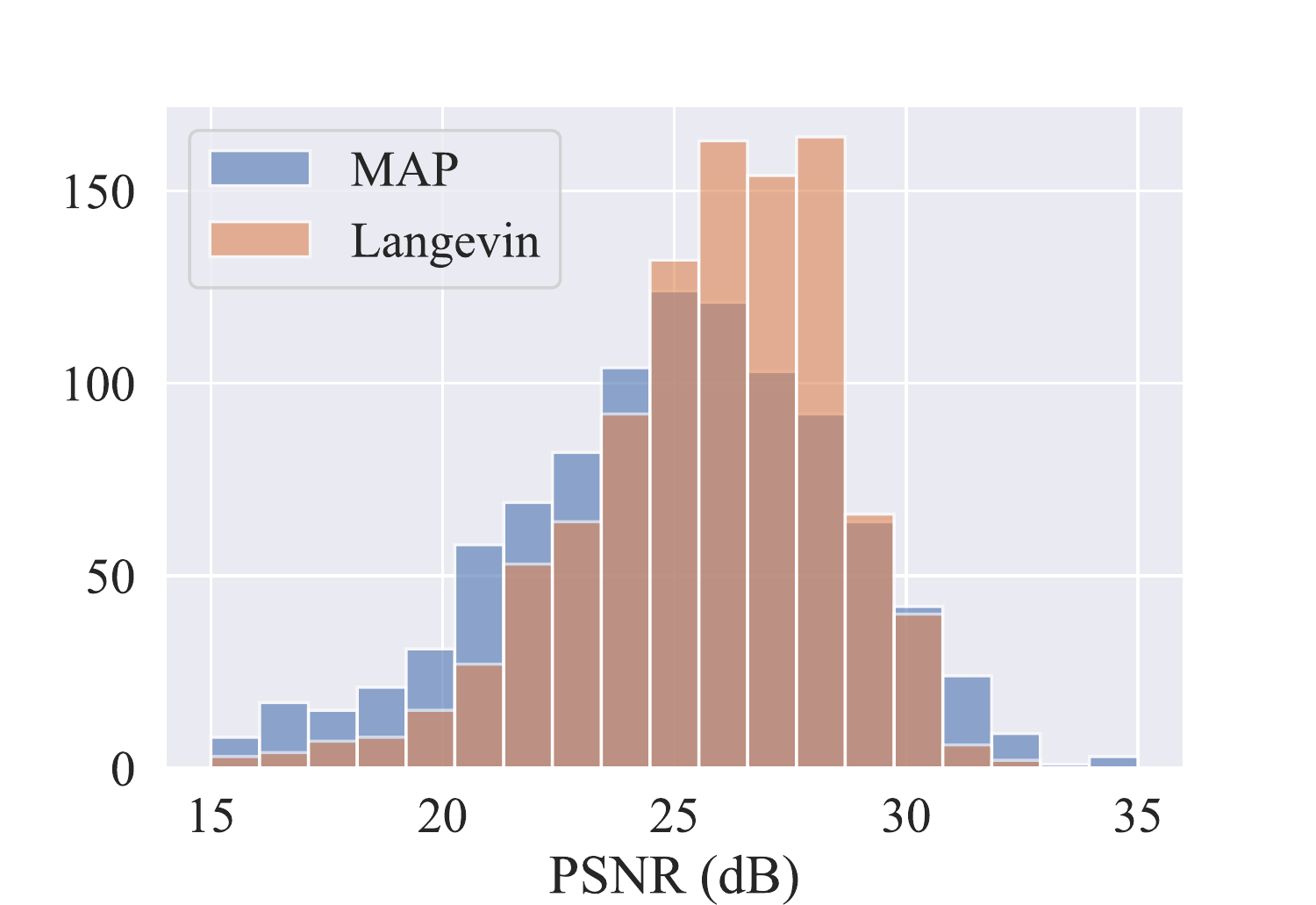}
\caption{}
\end{subfigure}
\begin{subfigure}[b]{0.45\textwidth}
\includegraphics[width=\textwidth]{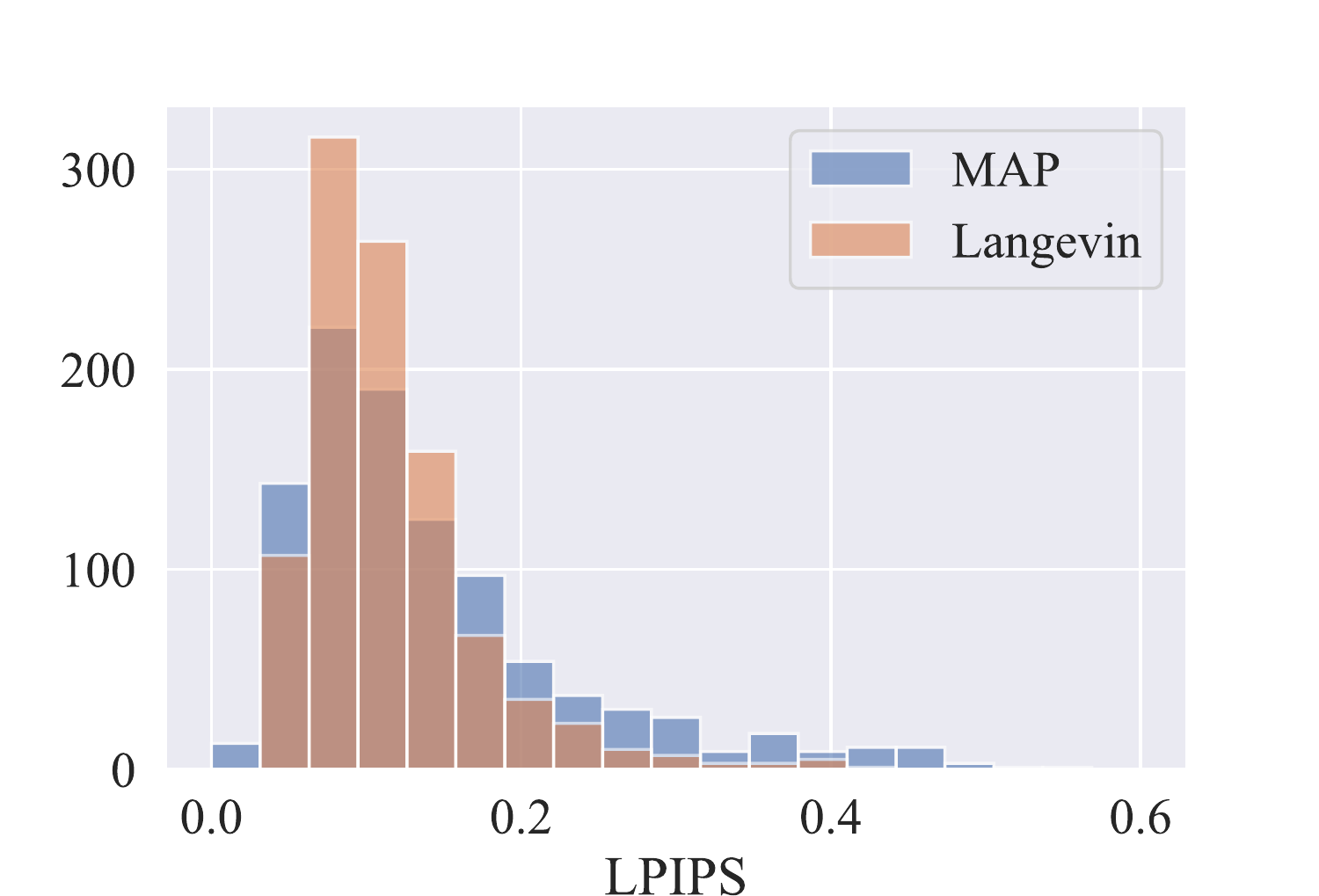}
\caption{}
\end{subfigure}
\caption{Histograms of blind image deblurring results on FFHQ ($256 \times 256$) dataset obtained from different update strategies for $\bm{\varphi}$. \textbf{MAP}: The linear operator's parameters are updated by MAP estimation, \textbf{Langevin}: GibbsDDRM, Proposed.}
\label{fig:histograms_MAP_Langevin}
\end{figure}

\begin{table}[htbp]
\begin{minipage}{0.32\linewidth}
  \centering
  \begin{tabular}{c|lll}
\hline
$\eta$ \textbackslash{} $\eta_b$ & \multicolumn{1}{c}{0.7} & \multicolumn{1}{c}{0.8} & \multicolumn{1}{c}{0.9} \\ \hline
0.7                              & 40.57                   & 39.43                   & 38.70                   \\
0.8                              & 40.59                   & 39.28                   & \textbf{38.51}          \\
0.9                              & 42.00                   & 40.68                   & 39.50                   \\ \hline
\end{tabular}
  \subcaption{FID ($\downarrow$)}
  \label{tab:table1}
\end{minipage}
\hfill
\begin{minipage}{0.32\linewidth}
  \centering
  \begin{tabular}{c|lll}
\hline
$\eta$ \textbackslash{} $\eta_b$ & \multicolumn{1}{c}{0.7} & \multicolumn{1}{c}{0.8} & \multicolumn{1}{c}{0.9} \\ \hline
0.7                              & 25.38                   & 25.65                   & 25.78                   \\
0.8                              & 25.39                   & 25.64                   & 25.78                   \\
0.9                              & 25.36                   & 25.62                   & \textbf{25.81}                   \\ \hline
\end{tabular}
  \subcaption{PSNR($\uparrow$)}
  \label{tab:table2}
\end{minipage}
\hfill
\begin{minipage}{0.32\linewidth}
  \centering
  \begin{tabular}{c|lll}
\hline
$\eta$ \textbackslash{} $\eta_b$ & \multicolumn{1}{c}{0.7} & \multicolumn{1}{c}{0.8} & \multicolumn{1}{c}{0.9} \\ \hline
0.7                              & 0.125                   & 0.118                   & \textbf{0.115}                   \\
0.8                              & 0.125                   & 0.119                   & \textbf{0.115}                   \\
0.9                              & 0.130                   & 0.123                   & 0.118                   \\ \hline
\end{tabular}
  \subcaption{LPIPS($\downarrow$)}
  \label{tab:table3}
\end{minipage}
\caption{Relationship between hyperparameters and evaluation metrics on FFHQ ($256 \times 256$) dataset. \textbf{Bold}: Best.}
\label{tab:relationship_etas}
\end{table}
\paragraph{Investigation of sampling methods of $\bm{\varphi}$.}
In GibbsDDRM, $\bm{\varphi}$ is sampled by Langevin dynamics using the estimated score in~\eqref{eq:est_score_phi}. If no Gaussian noise is added in Eq.~\eqref{eq:phi_langevin}, the operation can be interpreted as a step of gradient descent method for maximum a posteriori (MAP) estimation of $\bm{\varphi}$, with $\log p(\bm{\varphi}|\mathbf{x}_{t:T}, \mathbf{y})$ as the likelihood function. Although this operation cannot be included in GibbsDDRM as it is not a sampling of $\bm{\varphi}$, we can consider updating $\bm{\varphi}$ using this procedure. This strategy is referred to as ``MAP" and the GibbsDDRM as ``Langevin." Figure~\ref{fig:histograms_MAP_Langevin} shows histograms of PSNR and LPIPS computed for the images (in total 1000-images) estimated by Langevin (GibbsDDRM) and by MAP in the blind image deblurring experiment on FFHQ ($256\times 256$) dataset. It can be seen that the MAP's histogram has a longer tail, indicating that while MAP can sometimes estimate images with high accuracy, it is less stable compared to Langevin. This suggests that Langevin sampling serves to stabilize the estimation of $\bm{\varphi}$. 

\begin{figure}[htb]
\centering
\centerline{\includegraphics[width=5.5in]{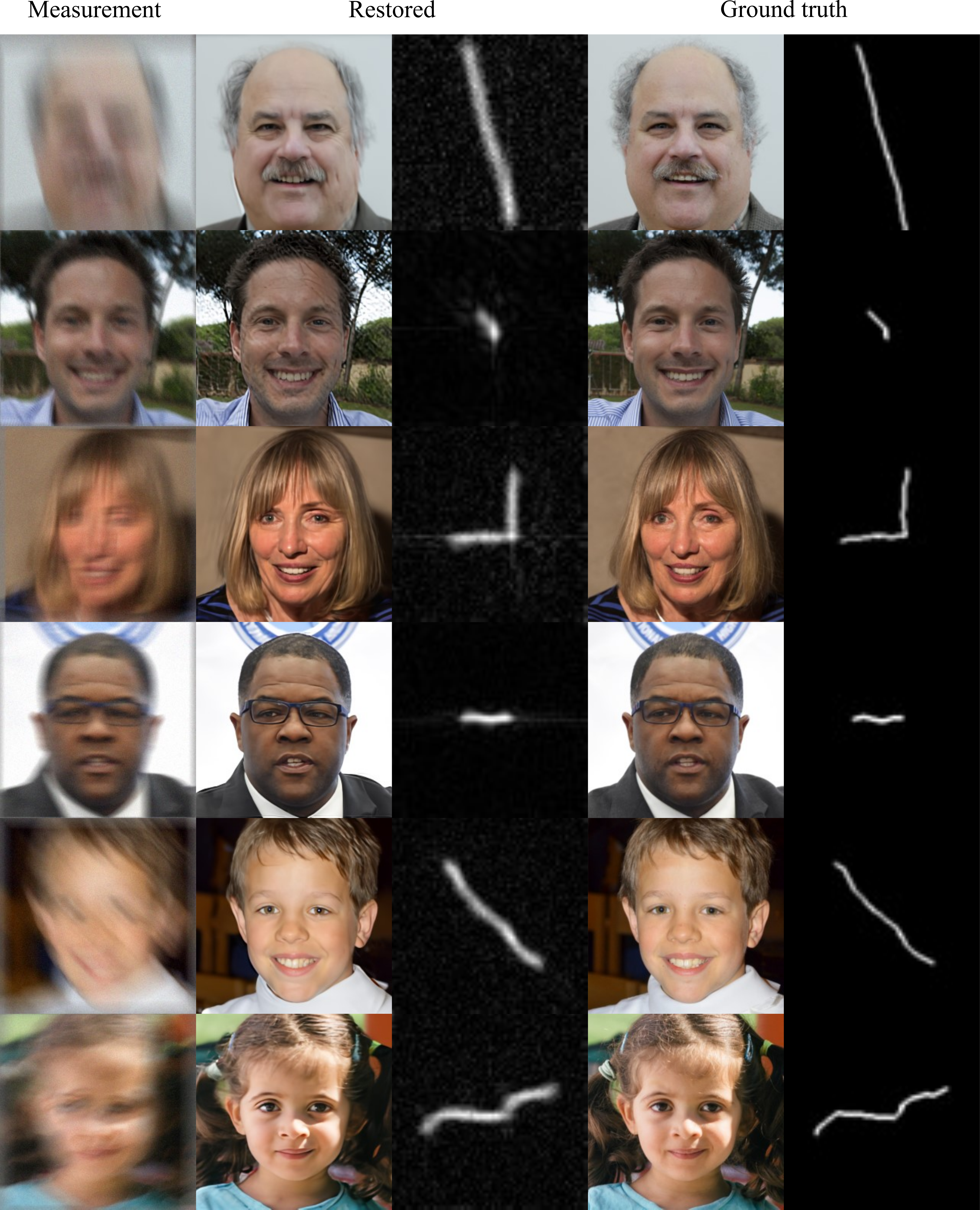}}
\caption{Blind image deblurring results obtained by GibbsDDRM on FFHQ ($256\times 256$) dataset.}
\label{fig:results_on_FFHQ}
\end{figure}

\begin{figure}[htb]
\centering
\centerline{\includegraphics[width=5.5in]{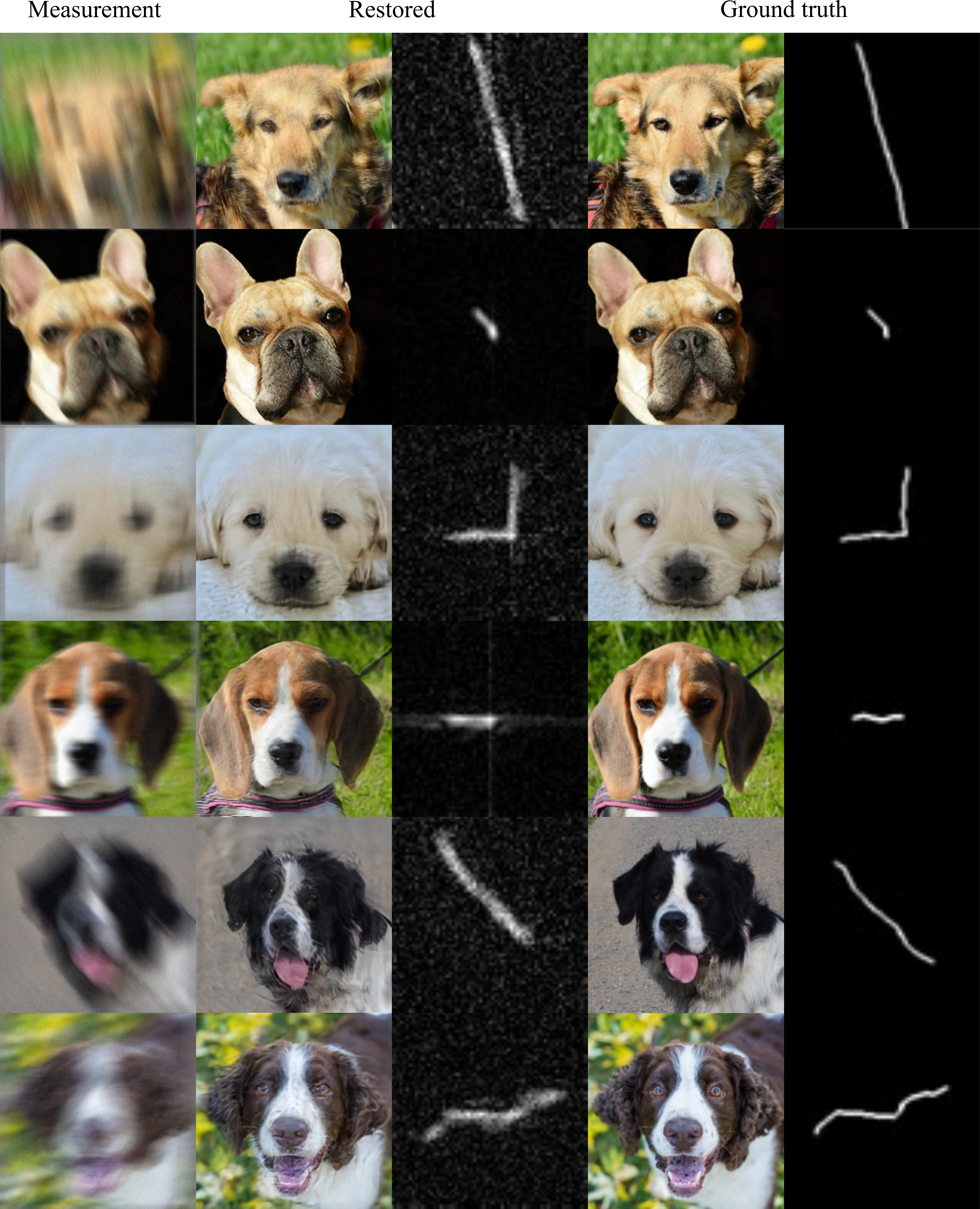}}
\caption{Blind image deblurring results obtained by GibbsDDRM on AFHQ ($256\times 256$) dataset.}
\label{fig:results_on_AFHQ}
\end{figure}

\paragraph{Additional figures}
We list additional qualitative results in Figs.~\ref{fig:results_on_FFHQ} and~\ref{fig:results_on_AFHQ} in order to see the details in the restored images and kernels.

\end{document}